\documentclass{article} 

\PassOptionsToPackage{numbers, compress, sort}{natbib}

\usepackage[final]{neurips_data_2023}

\usepackage{ulem}
\usepackage[utf8]{inputenc} 
\usepackage[T1]{fontenc}    
\usepackage[pagebackref=true,breaklinks=true,colorlinks,citecolor=citecolor, linkcolor=linkcolor]{hyperref}       
\usepackage{url}            
\usepackage{booktabs}       
\usepackage{amsfonts}       
\usepackage{nicefrac}       
\usepackage{microtype}      
\usepackage[table]{xcolor}  
\usepackage{bm}
\usepackage{enumitem}

\definecolor{citecolor}{HTML}{0071BC}
\definecolor{linkcolor}{HTML}{ED1C24}

\definecolor{deepfakecolor}{HTML}{CCEEFF}
\newcommand{\deepfake}[1]{\cellcolor{deepfakecolor}{#1}}

\usepackage{multirow}
\usepackage{booktabs}
\usepackage{subcaption}
\usepackage{amsmath,amsfonts,bm,amsthm,amssymb}
\usepackage{algorithm}
\usepackage{algorithmic}
\usepackage{graphics,graphicx,caption,float,color}
\usepackage{wrapfig}
\usepackage{bbm}
\usepackage{appendix}

\usepackage{listings}
\usepackage{xspace}
 
\usepackage{hhline}
\usepackage{multicol}
\usepackage{booktabs} 
\usepackage{makecell} 
\usepackage{bbding}
\usepackage{pifont}
\usepackage{ulem}

\definecolor{MyDarkBlue}{RGB}{0,0,180}
\definecolor{MyDarkGreen}{rgb}{0.02,0.6,0.02}
\definecolor{MyDarkRed}{rgb}{0.8,0.02,0.02}
\definecolor{MyDarkOrange}{rgb}{0.40,0.2,0.02}
\definecolor{MyPurple}{RGB}{111,0,255}
\definecolor{MyRed}{rgb}{1.0,0.0,0.0}
\definecolor{MyGold}{rgb}{0.75,0.6,0.12}
\definecolor{MyDarkgray}{rgb}{0.66, 0.66, 0.66}

\definecolor{ice}{RGB}{84,146,214}
\definecolor{fire}{RGB}{217,129,63}
\definecolor{baselinecolor}{gray}{.9}
\definecolor{poscolor}{RGB}{212,17,89}
\definecolor{negcolor}{RGB}{26,133,255}
\definecolor{bestcolor}{RGB}{238, 255, 238}
\newcolumntype{x}[1]{>{\centering\arraybackslash}p{#1pt}}
\newcolumntype{y}[1]{>{\raggedright\arraybackslash}p{#1pt}}
\newcolumntype{z}[1]{>{\raggedleft\arraybackslash}p{#1pt}}
\newcommand{\tablestyle}[2]{\setlength{\tabcolsep}{#1}\renewcommand{\arraystretch}{#2}\centering}

\graphicspath{{figures/}}

\title{\centering  

Seeing is not always believing: Benchmarking Human and Model Perception of AI-Generated Images

}

\newif\ifwithappendix
\withappendixtrue  %

\hypersetup{
        plainpages=false,
        colorlinks=true,              %
        linkcolor=blue,               %
        anchorcolor=blue,             %
        citecolor=blue,               %
        filecolor=blue,               %
        urlcolor=blue,                %
        pdfview=FitH,                 %
        pdfstartview=FitH,            %
        pdfpagelayout=SinglePage      %
}

\author{
Zeyu Lu$^{1,2,}$\thanks{Equal contribution. ~~\textsuperscript{\dag}Corresponding author: bailei@pjlab.org.cn.}
~~
Di Huang$^{2,3,*}$
~~
Lei Bai$^{2,*,\dagger}$
~~
Jingjing Qu $^{2}$
~~
Chengyue Wu $^{4}$
\\
\textbf{Xihui Liu}$^{4}$
~~
\textbf{Wanli Ouyang}$^{2}$
\\
$^{1}$ Shanghai Jiao Tong University
~~
$^{2}$ Shanghai Artificial Intelligence Laboratory
\\
$^{3}$ The University of Sydney
~~
$^{4}$ The University of Hong Kong
}

\ifwithappendix\else
\fi

\begin{document}

\maketitle


\begin{abstract}

Photos serve as a way for humans to record what they experience in their daily lives, and they are often regarded as trustworthy sources of information. However, there is a growing concern that the advancement of artificial intelligence (AI) technology may produce fake photos, which can create confusion and diminish trust in photographs.
This study aims to comprehensively evaluate agents for distinguishing state-of-the-art AI-generated visual content. 
Our study benchmarks both human capability and cutting-edge fake image detection AI algorithms, using a newly collected large-scale fake image dataset \textbf{Fake2M}. 
In our human perception evaluation, titled \textbf{HPBench}, we discovered that humans struggle significantly to distinguish real photos from AI-generated ones, with a misclassification rate of \textbf{38.7\%}. 
Along with this, we conduct the model capability of AI-Generated images detection evaluation \textbf{MPBench} and the top-performing model from MPBench achieves a \textbf{13\%} failure rate 
under the same setting used in the human evaluation.
We hope that our study can raise awareness of the potential risks of AI-generated images and 
facilitate further research to prevent the spread of false information.
More information can refer to \url{https://github.com/Inf-imagine/Sentry}. 

\end{abstract}
\begin{figure*}[!h]
    \vspace{-0.3cm}
    \centering
    \includegraphics[width=0.95\linewidth]{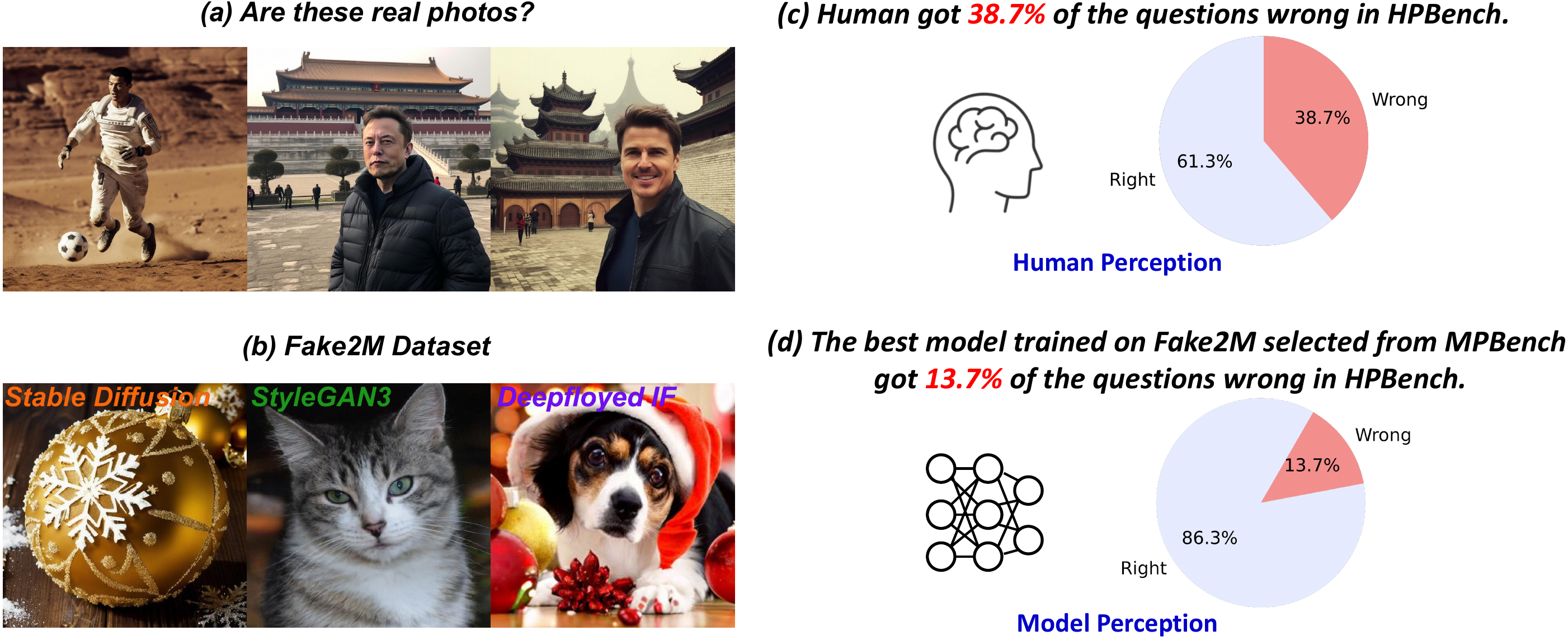}
    \caption{
        \textbf{\textit{Left:}}
        (a) Determining whether an image is real is a difficult problem.
        (b) We introduce a new, large-scale and diverse dataset, named "Fake2M".
        \textbf{\textit{Right:}}
        (c) The average human ability to distinguish between high-quality AI-generated images and real images is only 61\% in HPBench.
        (d) The highest performing model trained on the Fake2M dataset, selected from MPBench, achieves an accuracy of 87\% on HPBench.
    }
    \vspace{-0.3cm}
    \label{fake images}
\end{figure*}
\section{Introduction}



Photography, which captures images by recording light, has become an integral part of human society, serving as a vital medium for recording real visual information. 
Its diverse applications range from documenting historical events and scientific discoveries to immortalizing personal memories and artistic expressions. 
Since the advent of photography with Joseph Nicéphore Niépce's first photograph of photographs in 1826, the global collection of photographs has expanded to an estimated 12.4 trillion photos. 
This impressive number continues to rise with an annual increase of 1.72 trillion photos (10–14\% increasing rate every year)~\cite{photutorial}.

In contrast to traditional photography, AI-driven image generation harnesses neural networks to learn synthesis rules from extensive image datasets, offering a novel approach for creating high-quality yet fake visual content. By taking image descriptions or random noise as input, AI generates one or several images for users. Early methods employed GANs~\cite{Ian-2014-GAN, Andrew-2019-BigGAN, Han-2021-XMCGAN, Tero-2019-stylegan,Tero-2020-stylegan2,Tero-2020-stylegan3, Tero-2018-progan} as the generative architecture, while more recent diffusion-based techniques~\cite{Jonathan-2020-DDPM,Aditya-2022-DALLE2,Chitwan-2022-Imagen,Robin-2022-sd,song-2020-DDIM,Tero-2022-EDM,Nataniel-2022-dreambooth, Ben-2022-dreamfusion, Jonathan-2022-imagenvideo, Prafulla-2021-adm,Alexander-2021-iddpm,Zeyu-2023-HDAE,Patrick-2023-GEN1,song-2020-DDIM,yang-2021-score,Lvmin-2023-controlnet,Song-2021-scoremax,Song-2021-imporvescore,Yogesh-2022-edffi} have showcased improved diversity and generation quality. Therefore, AI-generated contents (AIGC) have gained popularity across various applications, such as AI-assisted design.

Following the rapid advancements in AI-driven image generation algorithms, AI is now capable of producing a wide array of image styles, including those that closely resemble real photographs. 
At the Sony World Photography Awards in April 2023, the winner Eldagsen refused the award after revealing that his creation is made using AI.
Organizers of the award said that Eldagsen had misled them about the extent of AI that would be involved~\cite{bbc}.
Consequently, a critical question arises: 
\textit{Can humans find a reliable solution for distinguishing whether an image is AI-generated?}
This inquiry fundamentally questions the reliability of image information for conveying truth. Utilizing generated images to convey false information can have significant social repercussions, often misleading people about nonexistent events or propagating incorrect ideas, as shown in Fig.~\ref{fake images}. 
A recent example involves the circulation of fake images depicting Trump's arrest~\cite{bbc2}, which garnered substantial social attention.


In order to tackle the challenge of discerning AI-generated images, we carried out an extensive study focusing on human capability and the proficiency of cutting-edge fake image detection AI algorithms. 
For human capability, we conduct a human evaluation involving participants from diverse backgrounds to determine their ability to distinguish real and fake images. Each participant is tasked with completing a test consisting of 100 randomized questions, where they must decide whether an image is real or generated by AI.
As for the evaluation of fake image detection models, existing datasets fell short in supporting this study due to their limited size and inclusion of outdated AI-generated images. 
Consequently, we assembled a novel dataset, called \textbf{Fake2M}, composed of state-of-the-art (SOTA) AI-generated images and real photographs sourced from the internet. 
Fake2M is a large-scale dataset housing over 2 million AI-generated images from three different models, tailored for training fake image detection algorithms.


For human evaluation of fake image detection task, our key finding underscores that state-of-the-art AI-generated images can indeed significantly deceive human perception. According to our results, participants achieved an average accuracy of only 61.3\%, implying a misclassification rate of 38.7\%, thus demonstrating the challenge they faced in accurately discerning real images from those generated by AI. Our study also reveals that humans are better at distinguishing AI-generated portrait images compared to other types of AI-generated images.

Turning to the model evaluation for the fake image detection task, our experiments yielded several key insights. 
Firstly, no single model consistently outperforms the others across all training datasets, suggesting that the optimal model is largely dependent on the specific dataset in use.
This underscores the importance of developing a robust model that can consistently achieve superior performance across all training datasets.
Secondly, our findings underscored the benefits of diverse training datasets, with models demonstrating improved overall accuracy when exposed to a broader range of visual styles and variations. 
Lastly, we observed variability in model accuracy when the same model trained on identical datasets was validated using different generation models. 
This demonstrates that variations in validation dataset generation may influence the performance of model.

In this study, we benchmark the ability of both human and cutting-edge fake image detection AI algorithms. While advancements in image synthesis have enabled individuals to create visually appealing images with quality approaching real photographs, our findings reveal the potential risks associated with using AIGC for spreading false information and misleading viewers. 
Our contributions can be listed as follows:
\begin{itemize}[leftmargin=*]
\item We introduce a new, large-scale dataset, named "\textbf{Fake2M}". 
To the best of our knowledge, this is the largest and most diverse fake image dataset, designed to stimulate and foster advancements in fake image detection research. 
\item We establish \textbf{HPBench}, a unique benchmark that comprehensively assesses the human capability to discern fake AI-generated images from real ones.
\item We establish \textbf{MPBench}, an extensive and comprehensive benchmark including 11 fake validation datasets, designed to rigorously assess the model capability for identifying fake images generated by the most advanced generative models currently available.

\end{itemize}

\section{Dataset Collection and Generation}

\subsection{Collect Data for Human Evaluation}


\begin{table*}[h]
\caption{
\textbf{Number of photographic images used in HPBench across eight categories.}
}
\centering
\renewcommand\arraystretch{1.4}
\resizebox{1\linewidth}{!}{
\begin{tabular}{l|cccccccc|c}
\toprule[1.2pt]
\textbf{Category}  & Multiperson & Landscape & Man & Woman & Record & Plant & Animal & Object& All \\ \midrule[1.1pt]
\textbf{Number of AI-Generated Images}                  & 10                          & 27                             & 17                       & 30                         & 15                          & 13                         & 29                          & 10         & 151  \\ 
\textbf{Number of Real Images}                   & 12                          & 26                             & 44                       & 49                         & 21                          & 18                         & 53                          & 21       &  244   \\ 
\bottomrule[1.2pt]
\end{tabular}
}
\label{data_info}
\end{table*}

We collect AI-generated images and a set of real photographs across eight categories: Multiperson, Landscape, Man, Woman, Record, Plant, Animal, and Object, as shown Tab.~\ref{data_info}. 

\noindent{\textbf{Collecting AI-generated photorealistic images.}}
Firstly, we utilize Midjourney-V5~\cite{midjourney}, the state-of-the-art image generation model, to create photorealistic images of the aforementioned eight categories. 
For each category, we employ diverse prompts to ensure maximum variation. 
We use specialized prompt suffixes such as "normal color, aesthetic, shocking, very detailed, photographic, 8K, HDR" to improve the authenticity of the images.
We notice that in real-world scenarios, people use AI to generate images with the intention of creating high-quality images without visual defects, and users will select the best image from multiple generated images.
Therefore, we employ the expertise of annotators to filter out low-quality AI-generated images that can be easily determined as fake photos at the first glance.


\noindent{\textbf{Collecting real photos.}}
We collect real photos from 500px~\cite{500px} and Google Images~\cite{google-images} by searching for photos with the same text prompts used for creating AI-generated images in the previous paragraph. 

\subsection{Collect Data for Model Evaluation}

\begin{table*}[!t]
\renewcommand\arraystretch{1.5}
\setlength\tabcolsep{2.pt}
\caption{
\textbf{Detailed information of the datasets used in MPBench.}
{\color[HTML]{CB0000} \textbf{R}} denotes the dataset consisting entirely of real images.
{\color[HTML]{036400} \textbf{F}} denotes the dataset consisting entirely of fake images.
{\color[HTML]{036400}{\ding{52}}} denotes existing datasets.
{\color[HTML]{CB0000}{\ding{56}}} denotes the datasets provided in this work.
"Diff" refers to diffusion model, "AR" refers to autoregressive model and "Unk." refers to unknown model.
}

\resizebox{1\linewidth}{!}{
\begin{tabular}{@{}l|ccccc|cccccccccccccc@{}}
\midrule[1.2pt]
\textbf{Dataset}                    &\rotatebox[origin=lb]{90} {\smash{\small CC3M-Train}}                           & \rotatebox[origin=lb]{90} {\smash{\small StyleGAN3-Train}}                       & \rotatebox[origin=lb]{90} {\smash{\small SD-V1.5Real-dpms-25 }}                  & \rotatebox[origin=lb]{90} {\smash{\small IF-V1.0-dpms++-25}}                     & \multicolumn{1}{c|}{\rotatebox[origin=lb]{90} {\smash{\small StyleGAN3}}}                            & \rotatebox[origin=lb]{90} {\smash{\small ImageNet-Test }}                       & \rotatebox[origin=lb]{90} {\smash{\small CelebA-HQ-Train  }}                    & \rotatebox[origin=lb]{90} {\smash{\small CC3M-Val }}                            & \rotatebox[origin=lb]{90} {\smash{\small SD-V2.1-dpms-25 }}                      & \rotatebox[origin=lb]{90} {\smash{\small SD-V1.5-dpms-25 }}                      & \rotatebox[origin=lb]{90} {\smash{\small SD-V1.5Real-dpms-25 }}                  & \rotatebox[origin=lb]{90} {\smash{\small IF-V1.0-dpms++-10  }}                   & \rotatebox[origin=lb]{90} {\smash{\small IF-V1.0-dpms++-25  }}                   & \rotatebox[origin=lb]{90} {\smash{\small IF-V1.0-dpms++-50  }}                   & \rotatebox[origin=lb]{90} {\smash{\small IF-V1.0-ddim-50   }}                   & \rotatebox[origin=lb]{90} {\smash{\small IF-V1.0-ddpms-50   }}                   & \rotatebox[origin=lb]{90} {\smash{\small Cogview2          }}                   & \rotatebox[origin=lb]{90} {\smash{\small Midjourney        }}                   & \rotatebox[origin=lb]{90} {\smash{\small StyleGan3         }}  \\ \midrule[1.2pt]
 & \multicolumn{5}{c|}{\textbf{Train}}& \multicolumn{14}{c}{\textbf{Validate}} \\
\multirow{-2}{*}{\textbf{Category}} & {\color[HTML]{CB0000} \textbf{R}} & {\color[HTML]{CB0000} \textbf{R}} & {\color[HTML]{036400} \textbf{F}} & {\color[HTML]{036400} \textbf{F}} & {\color[HTML]{036400} \textbf{F}} & {\color[HTML]{CB0000} \textbf{R}} & {\color[HTML]{CB0000} \textbf{R}} & {\color[HTML]{CB0000} \textbf{R}} & {\color[HTML]{036400} \textbf{F}} & {\color[HTML]{036400} \textbf{F}} & {\color[HTML]{036400} \textbf{F}} & {\color[HTML]{036400} \textbf{F}} & {\color[HTML]{036400} \textbf{F}} & {\color[HTML]{036400} \textbf{F}} & {\color[HTML]{036400} \textbf{F}} & {\color[HTML]{036400} \textbf{F}} & {\color[HTML]{036400} \textbf{F}} & {\color[HTML]{036400} \textbf{F}} & {\color[HTML]{036400} \textbf{F}} \\ \midrule[1.2pt]
\textbf{Generator}                  & -                                    & -                                    & Diff.                            & Diff.                            & GAN                                  & -                                    & -                                    & -                                    & Diff.                            & Diff.                            & Diff.                            & Diff.                            & Diff.                            & Diff.                            & Diff.                            & Diff.                            & AR                       & Unk.                              & GAN                                  \\
\textbf{Numbers}                    & 1M                                   & 87K                                & 1M                                   & 1M                                   & 87K                                & 100K                               & 24K                                & 15K                                & 15K                                & 15K                                & 15K                                & 15K                                & 15K                                & 15K                                & 15K                                & 15K                                & 22K                                & 5.5K                                 & 60K                                   \\
\textbf{This work}                       &           \color[HTML]{CB0000}{\ding{56}}                          &   \color[HTML]{CB0000}{\ding{56}}          &     \color[HTML]{036400}{\ding{52}}                                 &       \color[HTML]{036400}{\ding{52}}                               &    \color[HTML]{036400}{\ding{52}}                                  &     \color[HTML]{CB0000}{\ding{56}}                          &   \color[HTML]{CB0000}{\ding{56}}                           & \color[HTML]{CB0000}{\ding{56}}                           &  \color[HTML]{036400}{\ding{52}}                             &   \color[HTML]{036400}{\ding{52}}                         &  \color[HTML]{036400}{\ding{52}}                         & \color[HTML]{036400}{\ding{52}}                               &  \color[HTML]{036400}{\ding{52}}                          & \color[HTML]{036400}{\ding{52}}                             &          \color[HTML]{036400}{\ding{52}}               &  \color[HTML]{036400}{\ding{52}}                         &\color[HTML]{036400}{\ding{52}}                                & \color[HTML]{036400}{\ding{52}}                            & \color[HTML]{036400}{\ding{52}}                           \\ \bottomrule[1.2pt]

\end{tabular}
}

\label{model_dataset}
\end{table*}

Our objective is to investigate whether models can distinguish if an image is AI-generated or not. 
We constructed 3 training fake datasets with about 2M images, named \textbf{Fake2M}, and 11 validation fake datasets with about 257K images using different latest modern generative models, which contain the SOTA Diffusion models (Stable Diffusion~\cite{Robin-2022-sd}, IF~\cite{if}), the SOTA GAN model (StyleGAN3~\cite{Tero-2020-stylegan3}), the SOTA autoregressive model (CogView2~\cite{Ming-2021-CogView}), and the SOTA generative model (Midjounrey~\cite{midjourney}), as shown in Tab.~\ref{model_dataset}.
We describe the details of our datasets in the following subsections.

\noindent{\textbf{Collecting training datasets.}}
For the text-to-image generation model, we use the first 1M captions from CC3M to generate the corresponding fake images.
For the class conditional generation model, we generate the fake images directly.
We describe the specific settings of the 3 training fake datasets in Tab~\ref{model_dataset}, as follows:
\textbf{(1) "SD-V1.5Real-dpms-25":} Stable Diffusion v1.5 Realistic Vision V2.0~\cite{sdv1.5realv2.0} (the top popular models in CIVITAI) with DPM-Solver 25 steps to generate the corresponding fake images.
\textbf{(2) "IF-V1.0-dpms++-25":} IF v1.0~\cite{if} with DPM-Solver++~\cite{Cheng-2022-DPMSolver++} 25 steps to generate the corresponding fake images.
\textbf{(3) "StyleGAN3":} To match the training datasets domain in StyleGAN3~\cite{Tero-2020-stylegan3} and to maximize the diversity of the model, we use StyleGAN3-t-ffhq to generate the 35K fake images, StyleGAN3-r-ffhq to generate the 35K fake images, StyleGAN3-t-metfaces to generate the 650 fake images, StyleGAN3-r-metfaces to generate the 650 fake images, stylegan3-t-afhqv2 to generate the 8K fake images and stylegan3-r-afhqv2 to generate the 8K fake images.

In order to align with the domain of former fake datasets, we used the corresponding real images datasets and the specific settings of the 2 training real datasets in Tab~\ref{model_dataset}, as follows: 
\textbf{(1) "CC3M-Train":}
To conform with the former fake datasets, we use the corresponding real images from CC3M~\cite{Piyush-2018-cc3m}.
\textbf{(2) "StyleGAN3-Train":}
To match the former fake dataset, we use the StyleGAN3 training datasets: FFHQ~\cite{Tero-2019-stylegan}, AFHQv2~\cite{choi-2020-starganv2} and MetFaces~\cite{Tero-2020-stylegan3}.

\noindent{\textbf{Collecting validation datasets.}}
For the text-to-image generation model, we use the whole 15K captions from CC3M validation dataset to generate the corresponding fake images.
For the class conditional generation model, we generate fake images directly.
For Midjourney, we crawled 5.5K images from the community as a validation set.
We describe the specific settings of the 11 validation fake datasets in Tab~\ref{model_dataset}, as follows:
\textbf{(1) "SD-V2.1-dpms-25":} Stable Diffusion v2.1 with DPM-Solver~\cite{Cheng-2022-DPMSolver} 25 steps to generate the corresponding fake images.
\textbf{(2) "SD-V1.5-dpms-25":} Stable Diffusion v1.5 with DPM-Solver 25 steps to generate the corresponding fake images.
\textbf{(3) "SD-V1.5Real-dpms-25":} Stable Diffusion v1.5 Realistic Vision V2.0 with DPM-Solver 25 steps to generate the corresponding fake images.
\textbf{(4) "IF-V1.0-dpms++-10":} IF v1.0 with DPM-Solver++ 10 steps to generate the corresponding fake images.
\textbf{(5) "IF-V1.0-dpms++-25":} IF v1.0 with DPM-Solver++ 25 steps to generate the corresponding fake images.
\textbf{(6) "IF-V1.0-dpms++-50":} IF v1.0 with DPM-Solver++ 50 steps to generate the corresponding fake images.
\textbf{(7) "IF-V1.0-ddim-50":} IF v1.0 with DDIM~\cite{song-2020-DDIM} 50 steps to generate the corresponding fake images.
\textbf{(8) "IF-V1.0-ddpm-50":} IF v1.0 with DDPM~\cite{Jonathan-2020-DDPM} 50 steps to generate the corresponding fake images.
\textbf{(9) "Cogview2":} Cogview2 to generate the corresponding fake images.
\textbf{(10) "Midjourney":} We crawl 5K images from the community as a validation set.
\textbf{(11) "StyleGan3":} To maximize the diversity of the model, we generate 10K images each using StyleGAN3-t-ffhq, StyleGAN3-r-ffhq, StyleGAN3-t-metfaces, StyleGAN3-r-metfaces, stylegan3-t-afhqv2 and stylegan3-r-afhqv2, for a total of 60K images.

We use 3 common real datasets as our validation dataset and the specific settings as follows:
\textbf{(1) "ImageNet-Test":} The test dataset of ImageNet1k~\cite{jia-2009-imagenet}.
\textbf{(2) "CelebA-HQ-Train":} The training dataset of CelebA-HQ~\cite{Lee-2020-CelebAHQ}.
\textbf{(3) "CC3M-Val":} We use the validation dataset of CC3M as our validation dataset.

\subsection{Comparison with Other Datasets}

\begin{table*}[h]
\vspace{-0.2cm}
\caption{
\textbf{Comparison with other fake image datasets.}
"Content Type" means the type of the content in each dataset ("Face" means that the content in this dataset is mostly faces, such as FFHQ~\cite{Tero-2019-stylegan}. "Object" means that the content in this dataset is mainly composed of a limited number of objects, such as ImageNet~\cite{jia-2009-imagenet}. "General" means that the content in this dataset is general, not limited to some objects, faces or art, such as CC3M~\cite{Piyush-2018-cc3m}). "Generator Type" means the type of generator used in our dataset. "Public" means the dataset is publicly accessible. "Fake Image Number" represents the number of fake images provided by this dataset.
}
\centering
\renewcommand\arraystretch{1.4}
\resizebox{1\linewidth}{!}{

\begin{tabular}{l|ccccccc}
\toprule[1.3pt]
\multirow{2}{*}{Dataset} & \multirow{2}{*}{Content Type} & \multicolumn{3}{c}{Generator Type}       & \multirow{2}{*}{Public} &\multirow{2}{*}{Generator Number} & \multirow{2}{*}{Fake Image Number} \\ \cline{3-5}
                                       &                                & GAN        & Diffusion  & AutoRegressive &                         &                                     \\ \hline
FakeSpotter~\cite{Run-2020-fakespotter}                            & Face                           & \color[HTML]{036400}{\ding{52}}            & \color[HTML]{CB0000}{\ding{56}}          & \color[HTML]{CB0000}{\ding{56}}              & \color[HTML]{CB0000}{\ding{56}}                   & 7    & 5K                                  \\
DFFD~\cite{Hao-2020-DFFD}                                   & Face                           & \color[HTML]{036400}{\ding{52}}          & \color[HTML]{CB0000}{\ding{56}}          & \color[HTML]{CB0000}{\ding{56}}              & \color[HTML]{036400}{\ding{52}}               & 8        & 240K                                \\
APFDD~\cite{Apurva-2020-APFDD}                                  & Face                           & \color[HTML]{036400}{\ding{52}}          & \color[HTML]{CB0000}{\ding{56}}          & \color[HTML]{CB0000}{\ding{56}}              & \color[HTML]{CB0000}{\ding{56}}              & 1          & 5K                                  \\
ForgeryNet~\cite{Yinan-2021-ForgeryNet}                             & Face                           & \color[HTML]{036400}{\ding{52}}          & \color[HTML]{CB0000}{\ding{56}}          & \color[HTML]{CB0000}{\ding{56}}              & \color[HTML]{036400}{\ding{52}}         & 15              & 1.4M                                \\
DeepArt~\cite{Yabin-2023-DeepArt}                                & Art                            & \color[HTML]{CB0000}{\ding{56}}          & \color[HTML]{036400}{\ding{52}}          & \color[HTML]{CB0000}{\ding{56}}              & \color[HTML]{036400}{\ding{52}}         & 5              & 73K                                 \\
CNNSpot~\cite{Sheng-2020-GeneratedFake}                                & Object                         & \color[HTML]{036400}{\ding{52}}          & \color[HTML]{CB0000}{\ding{56}}          & \color[HTML]{CB0000}{\ding{56}}              & \color[HTML]{036400}{\ding{52}}         & 11              & 362K                                \\
IEEE VIP Cup~\cite{Luisa-2022-IEEEVIPCup}                           & Object                         & \color[HTML]{036400}{\ding{52}}          & \color[HTML]{036400}{\ding{52}}          & \color[HTML]{CB0000}{\ding{56}}                & \color[HTML]{CB0000}{\ding{56}}                 & 5      & 7K                                  \\
CiFAKE~\cite{CiFAKE-2023-Jordan}                                 & Object                         & \color[HTML]{CB0000}{\ding{56}}          & \color[HTML]{036400}{\ding{52}}          & \color[HTML]{CB0000}{\ding{56}}              & \color[HTML]{036400}{\ding{52}}            & 1           & 60K                                 \\
\textbf{Ours}                          & \textbf{General}            & \color[HTML]{036400}{\ding{52}} & \color[HTML]{036400}{\ding{52}} & \color[HTML]{036400}{\ding{52}}     & \color[HTML]{036400}{\ding{52}}  & \textbf{11}            & \textbf{2.3M}                       \\ 
\bottomrule[1.3pt]
\end{tabular}
}
\label{dataset_compare}
\end{table*}

As shown in Tab.~\ref{dataset_compare}, our dataset is
the largest and most diverse public general fake image dataset, designed to stimulate and foster advancements in
fake image detection research.

\section{HPBench: Human Perception of AI-Generated Images Evaluation}




Our objective is to investigate whether humans can distinguish if an image is AI-generated or not.
We collect a set of AI-generated photorealistic images and real photographs, and conduct a human evaluation  on the collected images to build \textbf{HPBench}. 
We describe the details of our data collection, human evaluation, and metrics for analyzing the results in the following subsections.


\subsection{Evaluation Setup}

\begin{table*}[!h]
\caption{\textbf{Number of participants across different backgrounds.} "w/ AIGC" refers to participants who have played with AIGC. "w/o AIGC" refers to participants who have not played with AIGC.}
\centering
\renewcommand\arraystretch{1.4}
\resizebox{0.63\linewidth}{!}{
\begin{tabular}{c|cc|cc|cc|c}
\toprule[1.2pt]
\multirow{3}{*}{\textbf{Category}}     & \multicolumn{2}{c|}{Gender} & \multicolumn{2}{c|}{Background} & \multicolumn{2}{c|}{Age} & \multirow{2}{*}{All}\\ 
                  & \multicolumn{1}{c}{Male} & \multicolumn{1}{c|}{Female}   & \multicolumn{1}{c}{w/ AIGC}       & \multicolumn{1}{c|}{w/o AIGC} &  \multicolumn{1}{c}{20$\sim$29} & \multicolumn{1}{c|}{30$\sim$45}&   \\ \midrule[1.1pt]
\multicolumn{1}{c|}{\textbf{Number}}  & \multicolumn{1}{c}{31} & 19& \multicolumn{1}{c}{27} & 23   & 42  & 8 &\multicolumn{1}{c}{50}                      \\
\bottomrule[1.2pt]
\end{tabular}
}
\label{user_info}
\end{table*} 

\noindent{\textbf{A high-quality fifty-participant human evaluation.}}
In order to ensure comprehensiveness, fairness, and quality of the evaluation, we recruit a total of 50 participants to participate in our human evaluation instead of using crowdsourcing and make efforts to ensure the diversity of the participants, as shown in Tab.~\ref{user_info}. 
Each participant is asked to complete a questionnaire consisting of 100 questions to determine whether the image is generated by AI or not without any time limit. 

\begin{table*}[h]
\centering
\caption{
\textbf{Human evaluation metrics across nine categories using all participant data.}
}
\renewcommand\arraystretch{1.4}
\resizebox{0.87\linewidth}{!}{
\begin{tabular}{l|ccccccccc}
\toprule[1.2pt]
\multirow{3}{*}{\textbf{Metric}} & \multicolumn{9}{c}{Category}                                              \\
                        & All & Multiperson & Landscape & Man & Woman & Record & Plant & Animal & Object \\ \midrule[1.2pt]
\textbf{Accuracy $\uparrow$}  &   0.6134  &  0.6750    &   0.5650    & 0.6433  &   0.6637   & 0.6233  & 0.5983  &   0.6133  &  0.5083   \\
\textbf{Precision $\uparrow$ } &  0.6278   &     0.7075   &  0.5657      &  0.6666 &  0.6765  &   0.6340  &  0.6213  &     0.6156   &   0.5112    \\
\textbf{Recall $\uparrow$ }    & 0.5577  &  0.5966     &   0.5714      &  0.5733  &  0.6275   &    0.5833    &  0.5033    &  0.6033    &  0.3800     \\
\textbf{FOR $\downarrow$}     &  0.3981   &   0.3487    & 0.4358  &  0.3742   &    0.3473  & 0.3858  &   0.4173  &  0.3888   & 0.4933 \\
\bottomrule[1.2pt]
\end{tabular}
}
\label{metric_all_persons}
\end{table*}

\noindent{\textbf{Evaluation metrics.}}
We employ four commonly used evaluation metrics to analyze our results and highlight their respective meanings in the context of our problem. We define positive samples as AI-generated images and negative samples as real images for our problem, and then calculate Accuracy, Precision, Recall, and False Omission Rate (FOR) in the context of our problem in Tab.~\ref{metric_all_persons}.






\vspace{-0.2cm}

\subsection{Results and Analysis}

\begin{wrapfigure}{!R}{0.32\textwidth}
    \centering
    \vspace{-0.7cm}
    \includegraphics[width=0.99\linewidth]{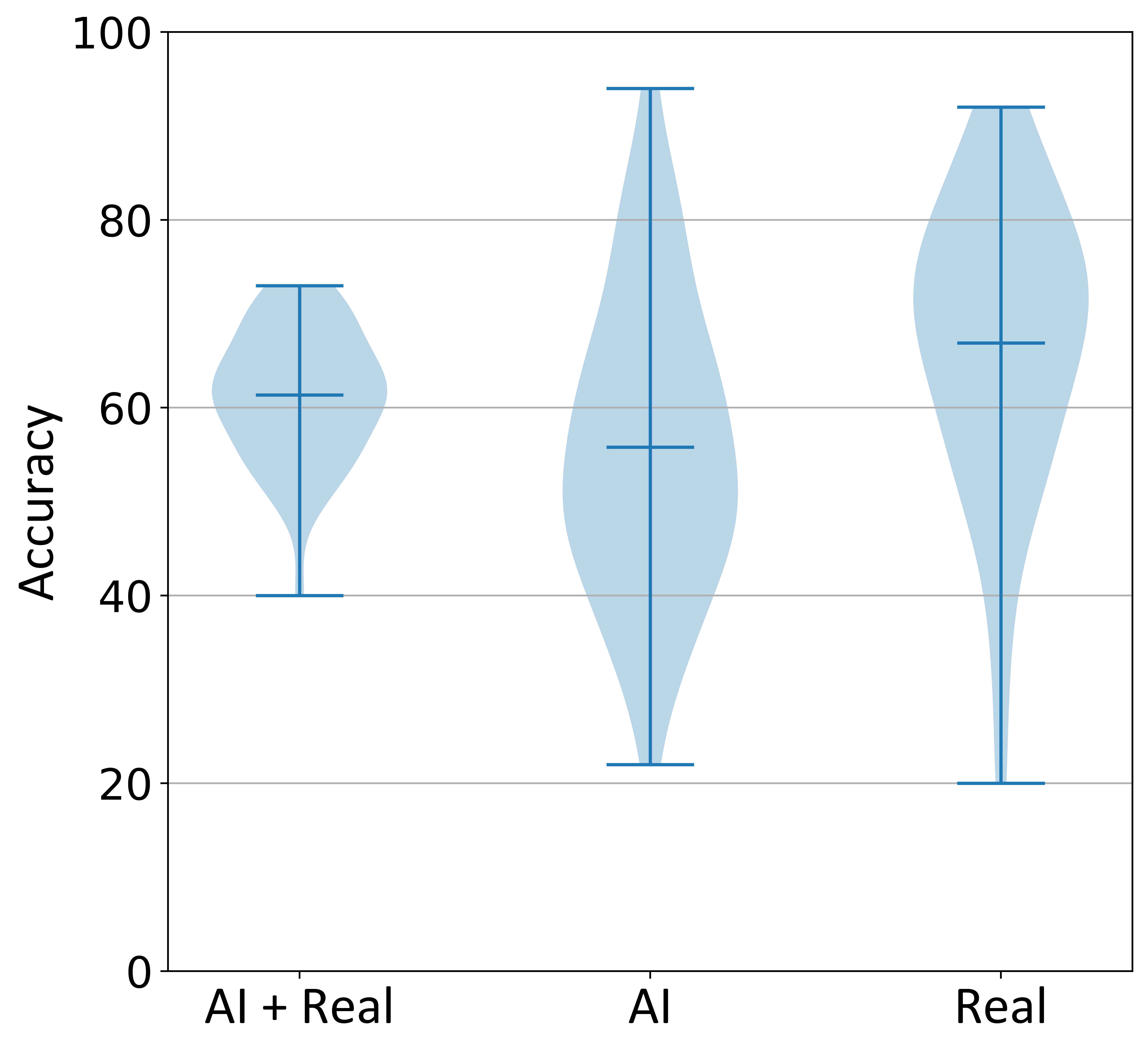}
    \vspace{-0.6cm}
    \caption{
        \textbf{Human evaluation score distribution  using all participant data.}
        "AI" represents only counting the AI-generated images. 
        "Real" represents only counting the real images. 
        "AI + Real" represents counting all the images. 
    }
    \vspace{-0.7cm}
    \label{all_categories_with_all_persons}
\end{wrapfigure}

\subsubsection{Overall Ability to Distinguish Real and AI-generated Images}

\paragraph{Results.}

The results are shown in Fig.~\ref{all_categories_with_all_persons}. Our study indicates that participants on average are able to correctly distinguish 61.3\% of the images, while 38.7\% of the images are misclassified. The highest-performing participant is able to correctly distinguish 73\% of the images, while the lowest-performing participant is only able to correctly distinguish 40\% of the images. These observations demonstrate that a combination of real and AI-generated fake images can easily deceive people. Moreover, the results reveal that humans have an average probability of 66.9\% of correctly identifying real photos from the internet, whereas for AI-generated images, people are more likely to be misled and incorrectly identify them as real with an average probability of 44.2\%.


\paragraph{Analysis.}
An intriguing observation from the above data is that even for real images collected from the Internet, participants are only able to correctly identify 66.9\% of the images that should have been correctly sorted out 100\% of the time. This finding demonstrates that AI-generated images not only convey incorrect information to humans but also erode people's trust in accurate information. 
Additionally, our study reveals that humans possess a superior ability to identify real photos than fake photos, which can be attributed to their lifetime experience.
Despite this, the difference between real image perception and AI-generated image perception is only 11.1\%, and we believe that future generative AI models could further narrow this gap.



\subsubsection{Distinguishing Abilities of Participants with Various Personal Backgrounds}

\begin{wrapfigure}{!R}{0.32\textwidth}
\vspace{-0.5cm}
    \centering
    \includegraphics[width=0.99\linewidth]{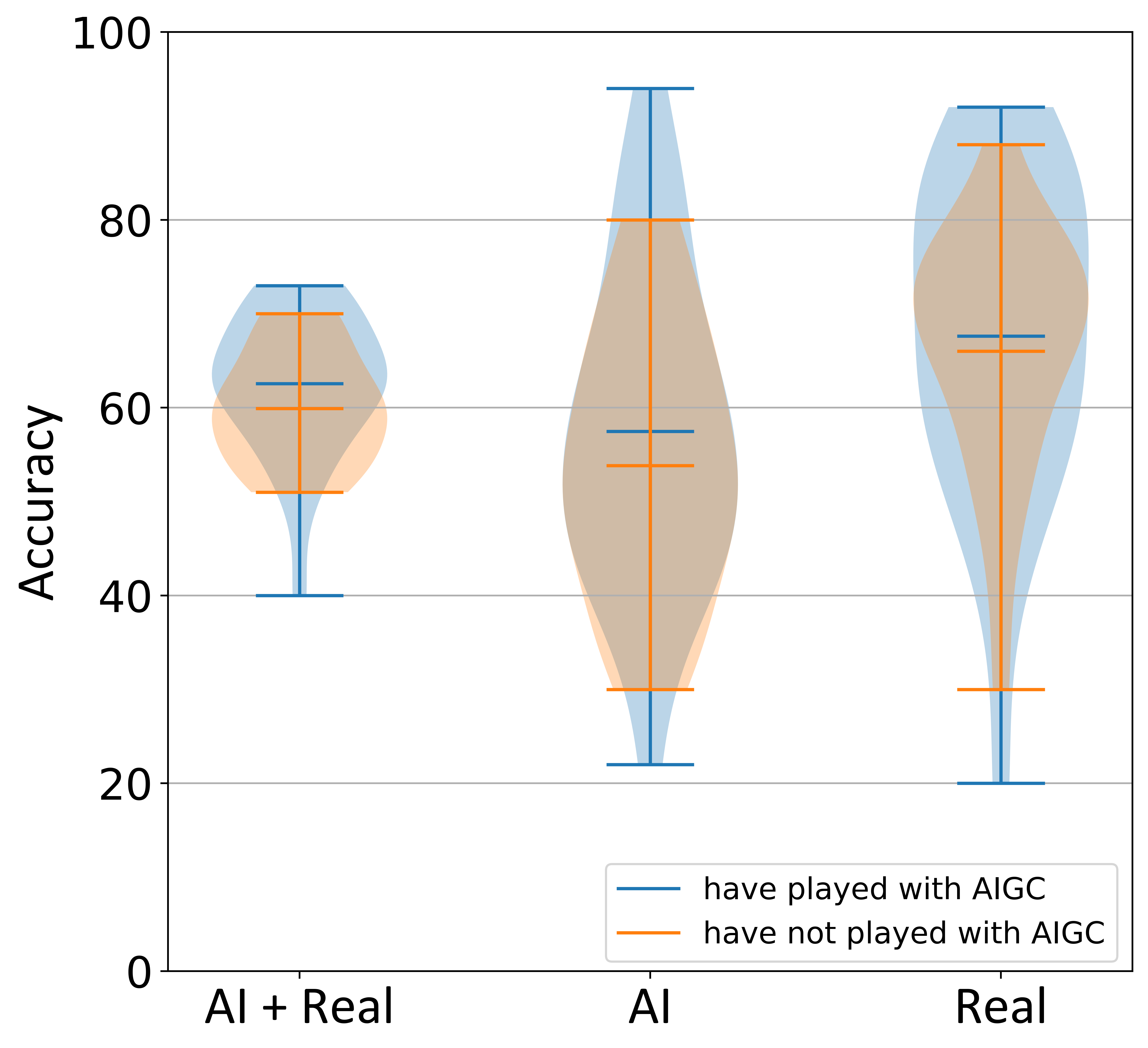}
    \caption{
        \textbf{Human evaluation score distribution calculated by data from participants with AIGC background and without AIGC background.}
    }
    \vspace{-0.3cm}
    \label{all categories with people have played with AIGC and have not played with AIGC}
\end{wrapfigure}

\paragraph{Results.}

We explore the effect of AIGC background (refers to participants who have played with AIGC) in our human evaluation. In Fig.~\ref{all categories with people have played with AIGC and have not played with AIGC}, we can see that individuals with AIGC background scored slightly better than those without AIGC background (+2.7\%). Interestingly, their AIGC background seem to have slight effects on their ability to identify real images (+0.7\%). However, when it comes to AI-generated images, participants with AIGC background performs significantly better, with a boost of 3.7\%. 

\vspace{-0.1cm}
\noindent{\textbf{Analysis.}}
Our research and analysis demonstrate that knowledge and experience with AIGC do not play a significant role in their ability to distinguish between real and AI-generated images. 
Specifically, when it comes to AI-generated images, participants with AIGC background have a slightly improved ability to distinguish between the two categories. 
Furthermore, this suggests that additional training and exposure to generative models may be beneficial for individuals by helping them make more informed decisions and avoid any potential risks from fake images.


\subsubsection{Distinguishability of different photo categories}

\begin{figure*}[!h]
    \centering
    \includegraphics[width=0.74\linewidth]{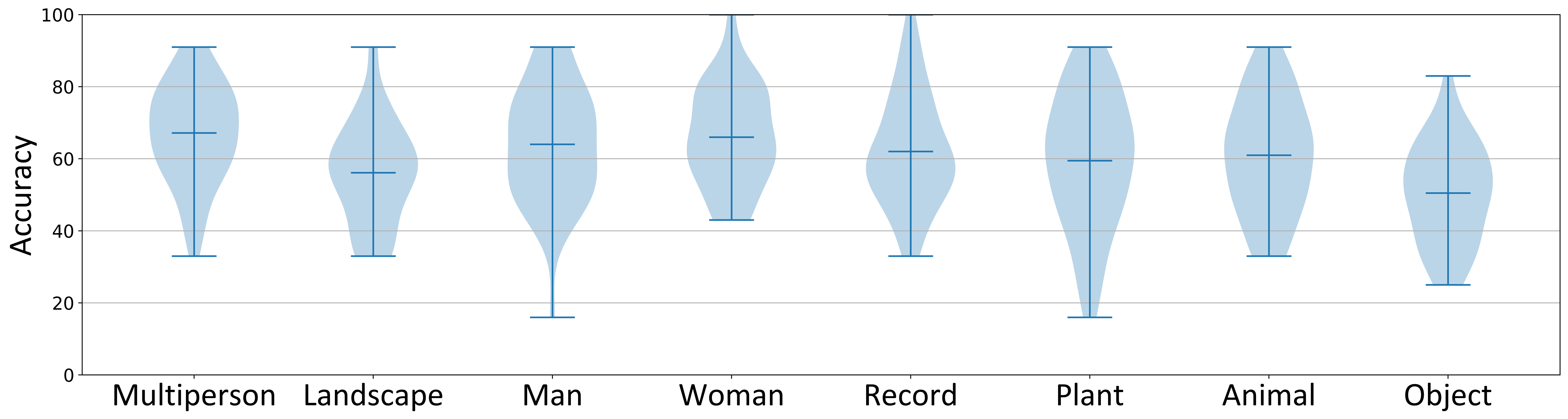}
    \caption{
        \textbf{Human evaluation score distribution across eight categories using all participant data.}
    }
    \label{All images with different categories for all persons}
\end{figure*}
\noindent{\textbf{Results.}}
We aim to investigate the difficulties for humans to distinguish real and fake images of different image categories.
The results are presented in Fig.~\ref{All images with different categories for all persons} and Tab.~\ref{metric_all_persons}. It is observed that the participants had varying ability levels in distinguishing real or fake images from different categories. For instance, the participants can distinguish real or fake images from the category \textit{Multiperson} with a relatively high accuracy rate of 67.5\%, whereas they can only correctly distinguish real or fake from the category \textit{Object} with a much lower accuracy rate of 50.8\%. Our results indicate a significant 16.7\% difference between the accuracy rates of the easiest and most challenging categories. These findings suggest that there may be significant differences in the way humans perceive different photo categories, which could have important implications for further research in this area.
It also indicates that the current AI-based generative models may be good at generating some categories but not so good at generating other categories.


\noindent{\textbf{Analysis.}}
The presented data offers valuable insights into the strengths and limitations of generative AI. The study indicates that current state-of-the-art AI image generation models excel in creating \textit{Object} images that are incredibly realistic. However, AI still struggles when it comes to generating human images. 
In Tab.~\ref{metric_all_persons}, \textit{Multiperson}, \textit{Man}, and \textit{Woman} categories are easier to be distinguished as real or fake images compared to others.
This phenomenon may be attributed to the fact that the human is more sensitive to images of humans, which is an essential aspect of our cognitive processing.

\begin{table*}[h]
\centering
\caption{\textbf{Statics of judgement criteria from participants correctly identifying fake images.}}
\renewcommand\arraystretch{1.4}
\resizebox{0.75\linewidth}{!}{
\begin{tabular}{l|cccccccc}
\toprule[1.2pt]
\multicolumn{1}{c|}{\textbf{Category}} & Detail & Smooth & Blur & Color & Shadow \& Light & Daub & Rationality & \noindent\textcolor{MyDarkgray}{Intuition} \\ \midrule[1.2pt]
\textbf{Number} & 332 & 205 & 142  & 122   & 95    & 59   & 57 & \noindent\textcolor{MyDarkgray}{169}  \\
\textbf{Percent} & 28\% & 17\% & 12\%  & 10\%   & 8\%    & 5\%   & 5\% & \noindent\textcolor{MyDarkgray}{14\%}  

\\ \bottomrule[1.2pt]
\end{tabular}
}
\label{judgement}
\end{table*}

\subsubsection{Results of the Judgment Criteria and Analysis of the AIGC Defects}


\noindent{\textbf{Results.}}
We predefined eight judgment criteria options based on our experience and the specific statistics of judgment criteria for correctly selecting the AI-generated images are shown in Tab.~\ref{judgement}.
The most common issues are "Detail problem" and "Smooth problem" with a high rate of 28\% and 17\%, respectively.
It is also worth noting that the proportion of participants who choose "\noindent\textcolor{MyDarkgray}{Intuition}" is about \noindent\textcolor{MyDarkgray}{14\%}, indicating that it is difficult for people to describe obvious defects in AI-generated images, even if they can successfully identify the AI-generated images.

\noindent{\textbf{Analysis.}}
Our experiment has revealed that even high-quality AI-generated images still exhibit various imperfections and shortcomings. Furthermore, the current SOTA image generation model has limited capability in generating fine details and often generates portraits that are overly smoothed.

\section{MPBench: Model Perception of AI-Generated Images  Evaluation}
Our objective is to investigate whether AI models can distinguish if an image is AI-generated or not.
We conduct a large and comprehensive model evaluation using \textbf{Fake2M} training dataset and 11 fake validation datasets to build \textbf{MPBench}.

\begin{table}[!h]
\renewcommand\arraystretch{1.3}
\setlength\tabcolsep{2pt}
\caption{
\textbf{Quantitative comparison of five models under four training dataset settings with fourteen validation datasets.}
"Diff" refers to diffusion model, "AR" refers to autoregressive model and "Unk." refers to unknown model.
{\color[HTML]{CB0000} \textbf{Real (R)}} denotes the dataset consisting entirely of real images.
{\color[HTML]{036400} \textbf{Fake (F)}} denotes the dataset consisting entirely of fake images.
}
\resizebox{1\linewidth}{!}{
\begin{tabular}{@{}l|c|cccc|cccccccccccc|c@{}}
\toprule[1.5pt]
&       & \rotatebox[origin=lb]{90} {\smash{\small ImageNet-Test}} & \rotatebox[origin=lb]{90} {\smash{\small CelebA-HQ-Train}} & \rotatebox[origin=lb]{90} {\smash{\small CC3M-Val}} & \cellcolor[HTML]{EFEFEF}\rotatebox[origin=lb]{90} {\smash{\small Average Acc.}}  & \rotatebox[origin=lb]{90} {\smash{\small SD-V2.1-dpm-25 }} & \rotatebox[origin=lb]{90} {\smash{\small SD-V1.5-dpm-25 }} & \rotatebox[origin=lb]{90} {\smash{\small SD-V1.5Real-dpm-25 }} & \rotatebox[origin=lb]{90} {\smash{\small IF-V1.0-dpm++-10 }} & \rotatebox[origin=lb]{90} {\smash{\small IF-V1.0-dpm++-25 }} & \rotatebox[origin=lb]{90} {\smash{\small IF-V1.0-dpm++-50 }} & \rotatebox[origin=lb]{90} {\smash{\small IF-V1.0-ddim-50 }} & \rotatebox[origin=lb]{90} {\smash{\small IF-V1.0-ddpm-50 }} & \rotatebox[origin=lb]{90} {\smash{\small Cogview2 }} & \rotatebox[origin=lb]{90} {\smash{\small Midjourney }} & \rotatebox[origin=lb]{90} {\smash{\small StyleGAN3 }} & \cellcolor[HTML]{EFEFEF} \rotatebox[origin=lb]{90} {\smash{\small Average Acc. }}& \cellcolor[HTML]{FFF8F8} \rotatebox[origin=lb]{90} {\smash{\small Total Average Acc. }} \\ 
\multicolumn{1}{l|}{}                        & \multicolumn{1}{c|}{}                               & \multicolumn{4}{c|}{{\color[HTML]{CB0000} \textbf{Real}}}                                                              & \multicolumn{12}{c|}{{\color[HTML]{036400} \textbf{Fake}}}                                                                                                                                                                                                                                                      &  {\color[HTML]{CB0000} \textbf{R}}+{\color[HTML]{036400} \textbf{F}}         \\
\multicolumn{1}{l|}{\multirow{1}{*}{\textbf{Model}}} & \multicolumn{1}{c|}{\multirow{1}{*}{\textbf{Training Dataset}}}        & -                    & -                    & -                    & \multicolumn{1}{c|}{-}    & Diff.                & Diff.                & Diff.                & Diff.                & Diff.                & Diff.                & Diff.                & Diff.                & AR                   & Unk.                 & GAN                  & \multicolumn{1}{c|}{-}     & -         \\ \midrule[1.5pt]
\multicolumn{1}{l|}{ConvNext-S(B+J 0.1)}     & \multicolumn{1}{c|}{}                               & 95.3                 & 99.9                 & 99.9                 & \multicolumn{1}{c|}{\cellcolor[HTML]{EFEFEF}98.3} & 48.6                 & 99.9                 & 85.1                 & 92.9                 & 57.6                 & 55.9                 & 41.1                 & 72.6                 & 55.7                 & 41.6                 & 37.4                 & \multicolumn{1}{c|}{\cellcolor[HTML]{EFEFEF}62.5}  & \cellcolor[HTML]{FFF8F8}70.2      \\
\multicolumn{1}{l|}{ConvNext-S(B+J 0.5)}     & \multicolumn{1}{c|}{}                               & 95.9                 & 99.9                 & 99.9                 & \multicolumn{1}{c|}{\cellcolor[HTML]{EFEFEF}\uline{98.5}} & 53.5                 & 100                  & 83.3                 & 91.1                 & 50.2                 & 49.6                 & 35.3                 & 66.9                 & 54.9                 & 44.7                 & 35.7                 & \multicolumn{1}{c|}{\cellcolor[HTML]{EFEFEF}60.4}  & \cellcolor[HTML]{FFF8F8}68.6      \\
\multicolumn{1}{l|}{ResNet50(B+J 0.1)}       & \multicolumn{1}{c|}{}                               & 93.0                 & 95.3                 & 95.6                 & \multicolumn{1}{c|}{\cellcolor[HTML]{EFEFEF}94.6} & 71.7                 & 71.8                 & 98.6                 & 57.2                 & 26.6                 & 29.0                 & 23.6                 & 47.9                 & 11.8                 & 40.3                 & 9.1                  & \multicolumn{1}{c|}{\cellcolor[HTML]{EFEFEF}44.3}  & \cellcolor[HTML]{FFF8F8}55.1      \\
\multicolumn{1}{l|}{ResNet50(B+J 0.5)}       & \multicolumn{1}{c|}{}                               & 93.2                 & 95.5                 & 95.8                 & \multicolumn{1}{c|}{\cellcolor[HTML]{EFEFEF}94.8} & 70.6                 & 70.4                 & 98.5                 & 52.1                 & 23.6                 & 26.1                 & 21.6                 & 46.2                 & 10.6                 & 36.6                 & 7.2                  & \multicolumn{1}{c|}{\cellcolor[HTML]{EFEFEF}42.1}  & \cellcolor[HTML]{FFF8F8}53.4      \\
\multicolumn{1}{l|}{CLIP-ViT-L(LC)}          & \multicolumn{1}{c|}{\multirow{-5}{*}{\makecell[c]{\textcolor{MyDarkgray}{Dataset Setting A:}\\ SD-V1.5Real-dpms-25 (1M)\\CC3M-Train (1M)}}}         & 49.6                 & 87.0                 & 75.4                 & \multicolumn{1}{c|}{\cellcolor[HTML]{EFEFEF}70.6} & 73.3                 & 86.6                 & 97.6                 & 93.9                 & 77.8                 & 71.4                 & 84.1                 & 90.9                 & 86.6                 & 88.7                 & 86.1                 & \multicolumn{1}{c|}{\cellcolor[HTML]{EFEFEF}\uline{85.1}}  & \cellcolor[HTML]{FFF8F8}\uline{82.0}      \\ \midrule
\multicolumn{1}{l|}{ConvNext-S(B+J 0.1)}     & \multicolumn{1}{c|}{}                               & 87.6                 & 99.9                 & 99.9                 & \multicolumn{1}{c|}{\cellcolor[HTML]{EFEFEF}\uline{95.8}} & 2.2                  & 34.5                 & 2.5                  & 99.7                 & 99.9                 & 99.9                 & 11.1                 & 66.4                 & 19.2                 & 8.9                  & 10.1                 & \multicolumn{1}{c|}{\cellcolor[HTML]{EFEFEF}41.3}  & \cellcolor[HTML]{FFF8F8}52.9      \\
\multicolumn{1}{l|}{ConvNext-S(B+J 0.5)}     & \multicolumn{1}{c|}{}                               & 87.8                 & 99.9                 & 99.9                 & \multicolumn{1}{c|}{\cellcolor[HTML]{EFEFEF}\uline{95.8}} & 3.9                  & 39.1                 & 3.9                  & 99.6                 & 99.9                 & 99.8                 & 18.5                 & 79.2                 & 25.8                 & 8.1                  & 8.0                  & \multicolumn{1}{c|}{\cellcolor[HTML]{EFEFEF}44.1}  & \cellcolor[HTML]{FFF8F8}55.2      \\
\multicolumn{1}{l|}{ResNet50(B+J 0.1)}       & \multicolumn{1}{c|}{}                               & 89.4                 & 95.8                 & 95.0                 & \multicolumn{1}{c|}{\cellcolor[HTML]{EFEFEF}93.4} & 37.5                 & 56.5                 & 20.0                 & 84.0                 & 95.6                 & 91.7                 & 39.7                 & 69.4                 & 45.3                 & 15.7                 & 8.8                  & \multicolumn{1}{c|}{\cellcolor[HTML]{EFEFEF}51.2}  & \cellcolor[HTML]{FFF8F8}60.3      \\
\multicolumn{1}{l|}{ResNet50(B+J 0.5)}       & \multicolumn{1}{c|}{}                               & 90.8                 & 95.6                 & 94.5                 & \multicolumn{1}{c|}{\cellcolor[HTML]{EFEFEF}93.6} & 41.6                 & 58.8                 & 21.7                 & 82.3                 & 95.2                 & 91.3                 & 47.0                 & 79.7                 & 56.1                 & 18.3                 & 6.8                  & \multicolumn{1}{c|}{\cellcolor[HTML]{EFEFEF}\uline{54.4}}  & \cellcolor[HTML]{FFF8F8}\uline{62.8}      \\
\multicolumn{1}{l|}{CLIP-ViT-L(LC)}          & \multicolumn{1}{c|}{\multirow{-5}{*}{\makecell[c]{\textcolor{MyDarkgray}{Dataset Setting B:}\\ IF-V1.0-dpms++-25 (1M)\\CC3M-Train (1M)}}}           & 81.5                 & 83.3                 & 93.0                 & \multicolumn{1}{c|}{\cellcolor[HTML]{EFEFEF}85.9} & 38.2                 & 18.5                 & 13.1                 & 80.4                 & 79.7                 & 70.9                 & 61.1                 & 77.7                 & 76.6                 & 33.7                 & 32.8                 & \multicolumn{1}{c|}{\cellcolor[HTML]{EFEFEF}52.9}  & \cellcolor[HTML]{FFF8F8}60.0      \\ \midrule
\multicolumn{1}{l|}{ConvNext-S(B+J 0.1)}     & \multicolumn{1}{c|}{}                               & 58.7                 & 81.0                 & 62.7                 & \multicolumn{1}{c|}{\cellcolor[HTML]{EFEFEF}67.4} & 44.9                 & 42.7                 & 39.4                 & 40.7                 & 42.3                 & 42.2                 & 35.7                 & 39.9                 & 53.2                 & 41.6                 & 69.9                 & \multicolumn{1}{c|}{\cellcolor[HTML]{EFEFEF}44.7}  & \cellcolor[HTML]{FFF8F8}49.6      \\
\multicolumn{1}{l|}{ConvNext-S(B+J 0.5)}     & \multicolumn{1}{c|}{}                               & 67.7                 & 92.6                 & 71.2                 & \multicolumn{1}{c|}{\cellcolor[HTML]{EFEFEF}\uline{77.1}} & 33.1                 & 33.7                 & 32.1                 & 33.6                 & 36.5                 & 34.2                 & 35.0                 & 35.5                 & 44.8                 & 28.5                 & 42.6                 & \multicolumn{1}{c|}{\cellcolor[HTML]{EFEFEF}35.4}  & \cellcolor[HTML]{FFF8F8}44.3      \\
\multicolumn{1}{l|}{ResNet50(B+J 0.1)}       & \multicolumn{1}{c|}{}                               & 31.5                 & 14.3                 & 39.1                 & \multicolumn{1}{c|}{\cellcolor[HTML]{EFEFEF}28.3} & 67.7                 & 62.7                 & 68.1                 & 65.3                 & 71.0                 & 67.0                 & 58.8                 & 60.9                 & 60.9                 & 81.9                 & 90.9                 & \multicolumn{1}{c|}{\cellcolor[HTML]{EFEFEF}68.6}  & \cellcolor[HTML]{FFF8F8}60.0      \\
\multicolumn{1}{l|}{ResNet50(B+J 0.5)}       & \multicolumn{1}{c|}{}                               & 70.6                 & 29.5                 & 63.4                 & \multicolumn{1}{c|}{\cellcolor[HTML]{EFEFEF}54.5} & 37.3                 & 34.9                 & 40.0                 & 43.3                 & 43.9                 & 41.9                 & 34.8                 & 38.4                 & 47.8                 & 58.1                 & 81.3                 & \multicolumn{1}{c|}{\cellcolor[HTML]{EFEFEF}45.6}  & \cellcolor[HTML]{FFF8F8}47.5      \\
\multicolumn{1}{l|}{CLIP-ViT-L(LC)}          & \multicolumn{1}{c|}{\multirow{-5}{*}{\makecell[c]{\textcolor{MyDarkgray}{Dataset Setting C:}\\ StyleGAN3 (87K)\\StyleGAN3-Train (87K)}}}    & 37.6                 & 85.9                 & 70.2                 & \multicolumn{1}{c|}{\cellcolor[HTML]{EFEFEF}64.5} & 85.7                 & 92.9                 & 94.8                 & 95.3                 & 89.5                 & 84.0                 & 87.5                 & 93.0                 & 84.6                 & 81.4                 & 61.9                 & \multicolumn{1}{c|}{\cellcolor[HTML]{EFEFEF}\textbf{\uline{86.4}}} & \cellcolor[HTML]{FFF8F8}\uline{81.7}      \\ \midrule
\multicolumn{1}{l|}{ConvNext-S(B+J 0.1)}     & \multicolumn{1}{c|}{}                               & 95.4                 & 99.9                 & 99.9                 & \multicolumn{1}{c|}{\cellcolor[HTML]{EFEFEF}98.4} & 39.9                 & 95.1                 & 99.9                 & 99.7                 & 99.8                 & 99.7                 & 43.1                 & 90.4                 & 48.3                 & 32.4                 & 99.8                 & \multicolumn{1}{c|}{\cellcolor[HTML]{EFEFEF}77.1}  & \cellcolor[HTML]{FFF8F8}81.6      \\
\multicolumn{1}{l|}{ConvNext-S(B+J 0.5)}     & \multicolumn{1}{c|}{}                               & 97.7                 & 99.9                 & 99.9                 & \multicolumn{1}{c|}{\cellcolor[HTML]{EFEFEF}\textbf{\uline{99.1}}} & 52.8                 & 92.9                 & 99.9                 & 99.5                 & 99.7                 & 99.3                 & 46.4                 & 91.7                 & 48.6                 & 35.0                 & 99.9                 & \multicolumn{1}{c|}{\cellcolor[HTML]{EFEFEF}78.7}  & \cellcolor[HTML]{FFF8F8}\textbf{\uline{83.0}}      \\
\multicolumn{1}{l|}{ResNet50(B+J 0.1)}       & \multicolumn{1}{c|}{}                               & 77.3                 & 93.6                 & 91.7                 & \multicolumn{1}{c|}{\cellcolor[HTML]{EFEFEF}87.5} & 83.9                 & 90.4                 & 96.8                 & 91.2                 & 90.6                 & 86.7                 & 52.3                 & 82.2                 & 56.4                 & 49.4                 & 80.8                 & \multicolumn{1}{c|}{\cellcolor[HTML]{EFEFEF}78.2}  & \cellcolor[HTML]{FFF8F8}80.2      \\
\multicolumn{1}{l|}{ResNet50(B+J 0.5)}       & \multicolumn{1}{c|}{}                               & 84.5                 & 94.7                 & 92.1                 & \multicolumn{1}{c|}{\cellcolor[HTML]{EFEFEF}90.4} & 84.4                 & 89.4                 & 96.1                 & 88.9                 & 89.4                 & 85.6                 & 52.4                 & 83.8                 & 55.4                 & 45.8                 & 69.5                 & \multicolumn{1}{c|}{\cellcolor[HTML]{EFEFEF}76.4}  & \cellcolor[HTML]{FFF8F8}79.4      \\
\multicolumn{1}{l|}{CLIP-ViT-L(LC)}          & \multicolumn{1}{c|}{\multirow{-5}{*}{\makecell[c]{\textcolor{MyDarkgray}{Dataset Setting D:}\\ SD-V1.5Real-dpms-25 (460K)\\IF-V1.0-dpms++-25 (460K)\\StyleGAN3 (87K)\\CC3M-Train (1M)\\StyleGAN3-Train (87K)}}} & 50.4                 & 96.0                 & 79.8                 & \multicolumn{1}{c|}{\cellcolor[HTML]{EFEFEF}75.4} & 66.9                 & 87.8                 & 93.6                 & 96.1                 & 85.9                 & 78.9                 & 84.9                 & 92.6                 & 87.4                 & 79.8                 & 62.8                 & \multicolumn{1}{c|}{\cellcolor[HTML]{EFEFEF}\uline{83.3}}  & \cellcolor[HTML]{FFF8F8}81.6      \\ \midrule[1.5pt]

\end{tabular}

}


\label{model_eval}
\end{table}

\subsection{Experiments Setup}
We compare the following SOTA methods: (1) Wang \textit{et al.}~\cite{Sheng-2020-GeneratedFake} proposed to finetune a classification network to give a real/fake decision for an image using \textit{Blur} and \textit{JPEG} augmentation with a pre-trained visual backbone, such as ResNet-50~\cite{Kaiming-2016-resnet} and ConvNext-S~\cite{Zhuang-2023-convnext}. (2) Ojha \textit{et al.}~\cite{Utkarsh-2023-UniFake} proposed to use a frozen CLIP-ViT-L~\cite{Alec-2021-CLIP} for backbone and train the last linear layer for the same task~\cite{Utkarsh-2023-UniFake}.

To ensure a fair comparison, we train each model using four different dataset settings, consisting of 1 million fake images and an equivalent number of real images:
(1) \textcolor{MyDarkgray}{Dataset Setting A}: "SD-V1.5Real-dpms-25" dataset with 1M fake images and "CC3M-Train" dataset with 1M real images. (2) \textcolor{MyDarkgray}{Dataset Setting B}: "IF-V1.0-dpms++-25" dataset with 1M fake images and "CC3M-Train" dataset with 1M real images. (3) \textcolor{MyDarkgray}{Dataset Setting C}: "StyleGAN3" dataset with 87K fake images and "StyleGAN3-Train" with 87K real images. (4) \textcolor{MyDarkgray}{Dataset Setting D}: "SD-V1.5Real-dpms-25, IF-V1.0-dpms++-25, StyleGAN3, CC3M-Train, StyleGAN3-Train" dataset with 460K fake images selected from "SD-V1.5Real-dpms-25" dataset, 460K fake images selected from "IF-V1.0-dpms++-25" dataset, 87K fake images from "StyleGAN3" dataset, 1M real images from "CC3M-Train" dataset and 87K real images from "StyleGAN3-Train" dataset.

We eval all the models using 11 validation datasets to build \textbf{MPBench}, as shown in Tab~\ref{model_eval}.
To clearly show the performance of the models under different validation datasets, we directly report the classification accuracy of each model.

\subsection{Results and Analysis}

\subsubsection{Comparative Analysis of Accuracy Across Various Models}

\noindent{\textbf{Results.}}
We conducted a systematic analysis of the mean accuracy achieved by different models, each trained using an identical dataset. Our findings are summarized in Tab.~\ref{model_eval}.
Notably, the leading performing models for fake image detection vary depending on the training dataset. 
For example, under \textcolor{MyDarkgray}{Dataset Setting A}, CLIP-ViT-L(LC) excels, whereas under \textcolor{MyDarkgray}{Dataset Setting B}, ResNet50 (B+J 0.5) outperforms others. Furthermore, CLIP-ViT-L(LC) achieves the highest fake image detection accuracies of 86.4\% and 83.3\% under \textcolor{MyDarkgray}{Dataset Setting C} and \textcolor{MyDarkgray}{Dataset Setting D} respectively.
A compelling finding from our study is the absence of a single model that consistently delivers superior performance across all dataset settings for fake image detection.
However, our experiments also suggest that ConvNext models are more adept at achieving higher average accuracies in detecting real images. In all four dataset settings, ConvNext consistently achieves top accuracies—98.5\%, 95.8\%, 77.1\%, and 99.1\% for \textcolor{MyDarkgray}{Dataset Setting A,B,C,D}, respectively.

\noindent{\textbf{Analysis.}}
Our observations highlight the need for models that perform optimally on both real and fake images. According to our study, while CLIP-ViT-L(LC) most often produces the best results for fake images, ConvNext outperforms CLIP-ViT-L(LC) in real image detection. However, real-world applications necessitate discerning between real and fake images. This underscores the need for future research to strike a balance between detection capabilities for both categories.

\subsubsection{Comparative Analysis of Accuracy Across Various Training Datasets}

\noindent{\textbf{Results.}}
We conduct an empirical analysis of the average accuracy of various models, each trained using different dataset configurations. The results, as presented in Tab.~\ref{model_eval}, indicate that \textcolor{MyDarkgray}{Dataset Setting D} generally yields superior model performance. ConvNext and ResNet50 models specifically show marked improvements when trained on \textcolor{MyDarkgray}{Dataset Setting D}. Meanwhile, CLIP-ViT-L (LC) demonstrates comparable performance across different dataset configurations, attaining accuracies of 82.0\% with \textcolor{MyDarkgray}{Dataset Setting A} and 81.6\% with \textcolor{MyDarkgray}{Dataset Setting D}. 

\noindent{\textbf{Analysis.}}
Our experimental findings suggest that the choice of training dataset significantly influences model performance in fake image detection tasks.
Notably, a diversified dataset like \textcolor{MyDarkgray}{Dataset Setting D}, which includes five distinct generative models, seems to enhance overall model accuracy. This improvement is likely due to the exposure to a broader spectrum of generative styles and variations that a diversified dataset offers. 
Additionally, our results highlight the aptitude of the proposed dataset, \textbf{Fake2M}, which features diverse data sources, for more generalized fake image detection.

\subsubsection{Comparative Analysis of Accuracy Across Various Validation Datasets}

\noindent{\textbf{Results.}}
In our experiment, we assessed the accuracy of a trained model across various validation datasets. As depicted in Tab.~\ref{model_eval}, the performance of a model varies significantly based on the generation models, sampling methods, and sampling steps used in the validation set. For instance, the ConvNext-S (B+J 0.5) model, when trained using \textcolor{MyDarkgray}{Dataset Setting D}, displayed a broad range of validation results for fake image detection, with accuracies spanning between 35.0\% and 99.9\%. These findings underscore the influence of validation set characteristics on a model's performance in the task of fake image detection.

\noindent{\textbf{Analysis.}}
In realistic applications, fake images can originate from a variety of generation models with a diverse range of hyperparameters. Consequently, an ideal fake image detection model should demonstrate consistent proficiency across this broad spectrum of generation settings. However, our experimental results show that existing models do not meet this requirement, indicating a need for the development of new approaches to handle the challenge of detecting fake images generated under varying settings.

\subsubsection{Evaluate the best model under the same setting used in HPBench.}
Based on its highest total average accuracy in MPBench,
we select ConvNext-S (B+J 0.5) with \textcolor{MyDarkgray}{Dataset Setting D} as the best model.
We then evaluate this model under the same setting in HPBench which consists of 50 real images and 50 fake images for testing and it achieves a 13\% failure rate.




\section{Related Work}

\noindent{\textbf{Image Generation Models.}}
State-of-the-art text-to-image synthesis approaches such as DALL·E 2~\cite{Aditya-2022-DALLE2}, Imagen~\cite{Chitwan-2022-Imagen}, Stable Diffusion~\cite{Robin-2022-sd}, IF~\cite{if}, and Midjourney~\cite{midjourney} have demonstrated the possibility of that generating high-quality, photorealistic images with diffusion-based generative models trained on large datasets~\cite{Christoph-2022-LAION5B}. Those models have surpassed previous GAN-based models~\cite{Ian-2014-GAN,Andrew-2019-BigGAN,Han-2021-XMCGAN,Tero-2019-stylegan} in both fidelity and diversity of generated images, without the instability and mode collapse issues that GANs are prone to.
In addition to diffusion models, other autoregressive models such as Make-A-Scene~\cite{Oran-2022-Make-A-Scene}, CogView~\cite{Ming-2021-CogView}, and Parti~\cite{Jiahui-2022-Parti} have also achieved amazing performance.

\noindent{\textbf{Fake Images Generation and Detection.}}
In recent years, there have been many works~\cite{Xu-2019-detectfake,Sheng-2020-GeneratedFake,Utkarsh-2023-UniFake,Lucy-2020-understand,Lakshmanan-2019-Detect,Vishal-2022-Proactive,Davide-2015-Splicebuster,Joel-2020-Frequency,Francesco-2018-Detection} exploring how to distinguish whether an image is AI-generated. 
These works focus on fake contents generated by GANs or small generation models~\cite{Ian-2014-GAN,Andrew-2019-BigGAN,Han-2021-XMCGAN,Tero-2019-stylegan}. 
Due to the limited quality of images generated by those methods, it is easy for humans to distinguish whether a photo is AI-generated or not. 
However, as the quality of generated images continues to improve with the advancement of recent generative models~\cite{Chitwan-2022-Imagen,midjourney,Robin-2022-sd,Aditya-2022-DALLE2}, it has become increasingly difficult for humans to identify whether an image is generated by AI or not. 
The need for detecting fake images has existed even before we had powerful image generators. 

\section{Conclusion}

In this study, we present a comprehensive evaluation of both human discernment and contemporary AI algorithms in detecting fake images. Our findings reveal that humans can be significantly deceived by current cutting-edge image generation models.
In contrast, AI fake image detection algorithms demonstrate a superior ability to distinguish authentic images from fakes.
Despite this, our research highlights that existing AI algorithms, with a considerable misclassification rate of 13\%, still face significant challenges. We anticipate that our proposed dataset, \textbf{Fake2M}, and our dual benchmarks, \textbf{HPBench} and \textbf{MPBench}, will invigorate further research in this area and assist researchers in crafting secure and reliable AI-generated content systems.
As we advance in this technological era, it is crucial to prioritize responsible creation and application of generative AI to ensure its benefits are harnessed positively for society.

\paragraph{Acknowledgments and Disclosure of Funding.}
This research is funded by Shanghai AI Laboratory.
This work is partially supported by the National Key R\&D Program of China(NO.2022ZD0160101).
We would like to thank many colleagues for useful discussions, suggestions, feedback, and advice, including: Peng Ye, Min Shi, Weiyun Wang, Yan Teng, Qianyu Guo, Kexin Huang, Yongqi Wang and Tianning Zhang.


\normalem
{\small
\bibliographystyle{plain}
\bibliography{ref}
}
\clearpage
\definecolor{Landscape}{RGB}{255,64,64}
\definecolor{Humans}{RGB}{0,215,240}
\definecolor{Plant}{RGB}{173,4,221}
\definecolor{Animal}{RGB}{0,25,212}
\definecolor{Record}{RGB}{255,166,79}
\definecolor{Object}{RGB}{250,2,252}
\definecolor{Man}{RGB}{244,242,7}
\definecolor{Woman}{RGB}{0, 255, 153}

\appendix


\section{Quick Test:  Can you identify which ones are AI-generated images?}


\begin{figure*}[!h]
    \centering    \includegraphics[width=0.9\linewidth]{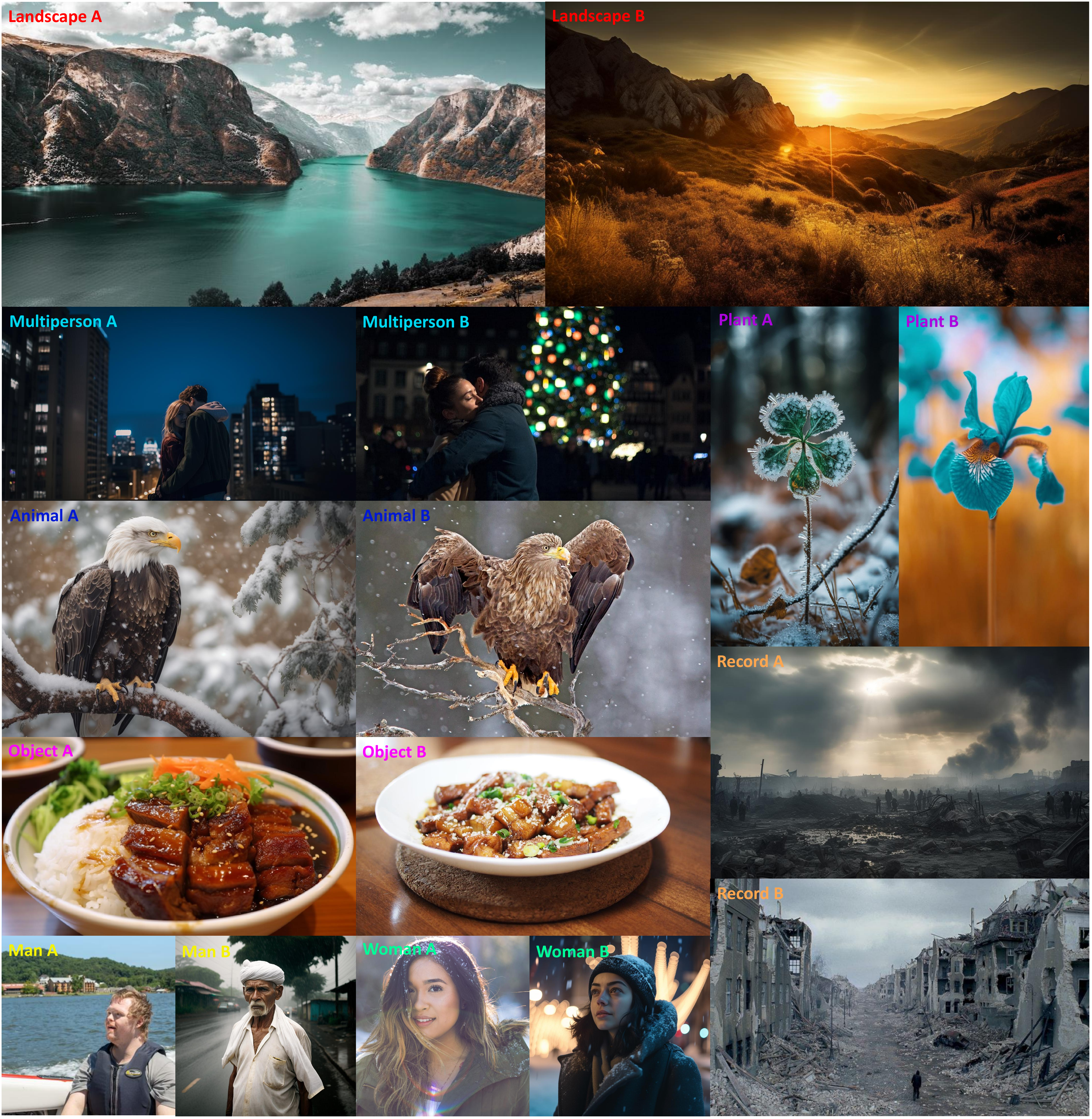}
    \vspace{-0.1cm}
    \caption{
        \textbf{A quick test:} \textit{Can you identify which ones are AI-generated images?}
    }
    \vspace{-0.3cm}
    \label{identify}
\end{figure*}

\paragraph{Answer of the Quick Test}
The AI-generated images of the Fig.~\ref{identify} are \textcolor{Landscape}{"Landscape B"}, \textcolor{Humans}{"Multiperson A"}, \textcolor{Plant}{"Plant A"}, \textcolor{Animal}{"Animal A"}, \textcolor{Record}{"Record A"}, \textcolor{Object}{"Object A"}, \textcolor{Man}{"Man B"}, \textcolor{Woman}{"Woman B"}, respectively.

\newpage

\section{Dataset}
\subsection{Dataset Configuration for Model Evaluation}

\begin{table}[!t]
\renewcommand\arraystretch{1.5}
\setlength\tabcolsep{2.pt}

\caption{
\textbf{Detailed information of the datasets used in MPBench.}
{\color[HTML]{CB0000} \textbf{R}} denotes the dataset consisting entirely of real images.
{\color[HTML]{036400} \textbf{F}} denotes the dataset consisting entirely of fake images.
{\color[HTML]{036400}{\ding{52}}} denotes existing datasets.
{\color[HTML]{CB0000}{\ding{56}}} denotes the datasets provided in this work.
"Diff" refers to diffusion model, "AR" refers to autoregressive model and "Unk." refers to unknown model.
"Resolution" refers to the resolution of the fake images in the dataset.
"Caption" refers to the caption used in text-to-image generation models to generate the corresponding dataset.
}

\resizebox{1\linewidth}{!}{
\begin{tabular}{@{}l|ccccc|cccccccccccccc@{}}
\midrule[1.2pt]

\textbf{Dataset}                   & \rotatebox[origin=lb]{90} {\smash{\small CC3M-Train}}            & \rotatebox[origin=lb]{90} {\smash{\small StyleGAN3-Train}}       & \rotatebox[origin=lb]{90} {\smash{\small SD-V1.5Real-dpms-25}}   & \rotatebox[origin=lb]{90} {\smash{\small IF-V1.0-dpms++-25}}     & \rotatebox[origin=lb]{90} {\smash{\small StyleGAN3}}              & \rotatebox[origin=lb]{90} {\smash{\small ImageNet-Test}}         & \rotatebox[origin=lb]{90} {\smash{\small CelebA-HQ-Train}}       & \rotatebox[origin=lb]{90} {\smash{\small CC3M-Val}}              & \rotatebox[origin=lb]{90} {\smash{\small SD-V2.1-dpms-25}}       & \rotatebox[origin=lb]{90} {\smash{\small SD-V1.5-dpms-25}}       & \rotatebox[origin=lb]{90} {\smash{\small SD-V1.5Real-dpms-25}}   & \rotatebox[origin=lb]{90} {\smash{\small IF-V1.0-dpms++-10}}     & \rotatebox[origin=lb]{90} {\smash{\small IF-V1.0-dpms++-25}}     & \rotatebox[origin=lb]{90} {\smash{\small IF-V1.0-dpms++-50}}     & \rotatebox[origin=lb]{90} {\smash{\small IF-V1.0-ddim-50}}       & \rotatebox[origin=lb]{90} {\smash{\small IF-V1.0-ddpms-50}}      & \rotatebox[origin=lb]{90} {\smash{\small Cogview2}}              & \rotatebox[origin=lb]{90} {\smash{\small Midjourney}}            & \rotatebox[origin=lb]{90} {\smash{\small StyleGan3}}             \\ \hline
\multirow{2}{*}{\textbf{Category}} & \multicolumn{5}{c|}{\textbf{Train}}                                                                                             & \multicolumn{14}{c}{\textbf{Validate}}                                                                                                                                                                                                                                                                                                                 \\
                          & {\color[HTML]{CB0000} \textbf{R}} & {\color[HTML]{CB0000} \textbf{R}} & {\color[HTML]{036400} \textbf{F}} & {\color[HTML]{036400} \textbf{F}} & {\color[HTML]{036400} \textbf{F}} & {\color[HTML]{CB0000} \textbf{R}} & {\color[HTML]{CB0000} \textbf{R}} & {\color[HTML]{CB0000} \textbf{R}} & {\color[HTML]{036400} \textbf{F}} & {\color[HTML]{036400} \textbf{F}} & {\color[HTML]{036400} \textbf{F}} & {\color[HTML]{036400} \textbf{F}} & {\color[HTML]{036400} \textbf{F}} & {\color[HTML]{036400} \textbf{F}} & {\color[HTML]{036400} \textbf{F}} & {\color[HTML]{036400} \textbf{F}} & {\color[HTML]{036400} \textbf{F}} & {\color[HTML]{036400} \textbf{F}} & {\color[HTML]{036400} \textbf{F}} \\ \hline
\textbf{Generator}                 & -                     & -                     & Diff.                 & Diff.                 & GAN                    & -                     & -                     & -                     & Diff.                 & Diff.                 & Diff.                 & Diff.                 & Diff.                 & Diff.                 & Diff.                 & Diff.                 & AR                    & Unk.                  & GAN                   \\
\textbf{Numbers}                   & 1M                    & 87K                   & 1M                    & 1M                    & 87K                    & 100K                  & 24K                   & 15K                   & 15K                   & 15K                   & 15K                   & 15K                   & 15K                   & 15K                   & 15K                   & 15K                   & 22K                   & 5.5K                  & 60K                   \\
\textbf{Resolution}                & -                     & -                     & 512               & 256               & {\small(>=)}512                      & -                     & -                     & -                     & 512               & 512               & 512               & 256               & 256               & 256               & 256               & 256               & 480               & {\small(>=)}640             & {\small(>=)}512             \\
\textbf{Caption}                   & -                     & -                     & CC3M-train(first 1M)          & CC3M-train(first 1M)          & -                      & -                     & -                     & -                     & CC3M-val      & CC3M-val      & CC3M-val      & CC3M-val      & CC3M-val      & CC3M-val      & CC3M-val      & CC3M-val      & CC3M-val      & -                     & -                     \\ 
\textbf{Seed}                   & -                     & -                     & 420          & 420          & -                      & -                     & -                     & -                     & 420      & 420      & 420      & 420      & 420      & 420      & 420     & 420      & -      & -                     & -                     \\ 
\textbf{CFG-Scale}                   & -                     & -                     & 7          & 7          & -                      & -                     & -                     & -                     & 7      & 7      & 7      & 7      & 7      & 7      & 7     & 7      & -      & -                     & -                     \\ 
\textbf{This work}                       &           \color[HTML]{CB0000}{\ding{56}}                          &   \color[HTML]{CB0000}{\ding{56}}          &     \color[HTML]{036400}{\ding{52}}                                 &       \color[HTML]{036400}{\ding{52}}                               &    \color[HTML]{036400}{\ding{52}}                                  &     \color[HTML]{CB0000}{\ding{56}}                          &   \color[HTML]{CB0000}{\ding{56}}                           & \color[HTML]{CB0000}{\ding{56}}                           &  \color[HTML]{036400}{\ding{52}}                             &   \color[HTML]{036400}{\ding{52}}                         &  \color[HTML]{036400}{\ding{52}}                         & \color[HTML]{036400}{\ding{52}}                               &  \color[HTML]{036400}{\ding{52}}                          & \color[HTML]{036400}{\ding{52}}                             &          \color[HTML]{036400}{\ding{52}}               &  \color[HTML]{036400}{\ding{52}}                         &\color[HTML]{036400}{\ding{52}}                                & \color[HTML]{036400}{\ding{52}}                            & \color[HTML]{036400}{\ding{52}}                           \\ \bottomrule[1.2pt]
\end{tabular}
}

\label{model_dataset_detail}
\end{table}
\begin{table*}[!t]
\caption{\textbf{Detailed information of the diffusion datasets used in MPBench.}}
\resizebox{1\linewidth}{!}{
\begin{tabular}{l|cc|cccccccc}
\toprule[1.3pt]
  \textbf{Category}              & \multicolumn{2}{c|}{\textbf{Train}}                                                                                                                                                            & \multicolumn{6}{c}{\textbf{Test}}                                                                                                                                                              \\ \hline

\textbf{Generators}      & \rotatebox[origin=lb]{90} {\smash{\small SD-V1.5Real-dpms-25}} & \rotatebox[origin=lb]{90} {\smash{\small IF-V1.0-dpms++-25}} & \rotatebox[origin=lb]{90} {\smash{\small SD-V2.1-dpms-25}} & \rotatebox[origin=lb]{90} {\smash{\small SD-V1.5-dpms-25}} & \rotatebox[origin=lb]{90} {\smash{\small SD-V1.5Real-dpms-25}} & \rotatebox[origin=lb]{90} {\smash{\small IF-V1.0-dpms++-10}} & \rotatebox[origin=lb]{90} {\smash{\small IF-V1.0-dpms++-25 }}& \rotatebox[origin=lb]{90} {\smash{\small IF-V1.0-dpms++-50}} & \rotatebox[origin=lb]{90} {\smash{\small IF-V1.0-ddim-50}} & \rotatebox[origin=lb]{90} {\smash{\small IF-V1.0-ddpms-50}}\\ \hline
\textbf{Total Numbers } & 1M & 1M & 15K & 15K & 15K & 15K & 15K & 15K & 15K & 15K \\ 
\textbf{Sampling Steps}         & 25                         & 25                         & 25                         & 25                         & 25 & 10 & 25 & 50 & 50 & 50 \\
\textbf{Sampling Methods}         & dpm-sovler~\cite{Cheng-2022-DPMSolver}                         & dpm-sovler++~\cite{Cheng-2022-DPMSolver++}                         & dpm-sovler~\cite{Cheng-2022-DPMSolver}                          & dpm-sovler~\cite{Cheng-2022-DPMSolver}                          & dpm-sovler~\cite{Cheng-2022-DPMSolver}  & dpm-sovler++~\cite{Cheng-2022-DPMSolver++} & dpm-sovler++~\cite{Cheng-2022-DPMSolver++} & dpm-sovler++~\cite{Cheng-2022-DPMSolver++} & ddim~\cite{song-2020-DDIM} & ddpm~\cite{Jonathan-2020-DDPM}s \\
\textbf{Seed}         & 420                         & 420                         & 420                         & 420                         & 420 & 420 & 420 & 420 & 420 & 420 \\
\textbf{CFG-Scale} & 7 & 7 & 7 & 7 & 7 & 7 & 7 & 7 & 7 & 7 \\
\textbf{Model} & Stable Diffusion v1.5 Realistic Version & IF v1.0 & Stable Diffusion v2.1 & Stable Diffusion v1.5 & Stable Diffusion v1.5 Realistic Version & IF v1.0 & IF v1.0 & IF v1.0 & IF v1.0 & IF v1.0
\\ \bottomrule[1.3pt]
\end{tabular}
}
\label{diffusion_detailed}
\end{table*}
\begin{table*}[!t]
\caption{\textbf{Detailed information of the StyleGAN3 datasets used in MPBench.}}
\resizebox{1\linewidth}{!}{
\begin{tabular}{l|cccccc|cccccc}
\toprule[1.3pt]
  \textbf{Category}              & \multicolumn{6}{c|}{\textbf{Train}}                                                                                                                                                            & \multicolumn{6}{c}{\textbf{Validate}}                                                                                                                                                              \\ \hline

\textbf{Generators}      & \rotatebox[origin=lb]{90} {\smash{\small stylegan3-r-ffhqu-1024x1024}} & \rotatebox[origin=lb]{90} {\smash{\small stylegan3-t-ffhqu-1024x1024}} & \rotatebox[origin=lb]{90} {\smash{\small stylegan3-r-afhqv2-512x512}} & \rotatebox[origin=lb]{90} {\smash{\small stylegan3-t-afhqv2-512x512}} & \rotatebox[origin=lb]{90} {\smash{\small stylegan3-r-metfaces-1024x1024}} & \rotatebox[origin=lb]{90} {\smash{\small stylegan3-t-metfaces-1024x1024}} & \rotatebox[origin=lb]{90} {\smash{\small stylegan3-r-ffhqu-1024x1024 }}& \rotatebox[origin=lb]{90} {\smash{\small stylegan3-t-ffhqu-1024x1024}} & \rotatebox[origin=lb]{90} {\smash{\small stylegan3-r-afhqv2-512x512}} & \rotatebox[origin=lb]{90} {\smash{\small stylegan3-t-afhqv2-512x512}} & \rotatebox[origin=lb]{90} {\smash{\small stylegan3-r-metfaces-1024x1024}} & \rotatebox[origin=lb]{90} {\smash{\small stylegan3-t-metfaces-1024x1024}} \\ \hline
\textbf{Total Numbers }  & \multicolumn{6}{c|}{87K}                                                                                                                                                              & \multicolumn{6}{c}{60K}                                                                                                                                                               \\ 
\textbf{Numbers}         & 35K                         & 35K                         & 8K                         & 8K                         & 0.65K                          & 0.65K                          & 10K                         & 10K                         & 10K                        & 10K                        & 10K                            & 10K                            \\ 
\textbf{Seeds}           & 10001$\sim$45000            & 10001$\sim$45000            & 10001$\sim$18000           & 10001$\sim$18000           & 10001$\sim$10800               & 10001$\sim$10800               & 1-10000                     & 1-10000                     & 1-10000                    & 1-10000                    & 1-10000                        & 1-10000                        \\ 
\textbf{Matched Dataset} & \multicolumn{2}{c}{FFHQ (70K)~\cite{Tero-2019-stylegan}}                            & \multicolumn{2}{c}{AFHQv2 (16K)~\cite{choi-2020-starganv2}}                        & \multicolumn{2}{c|}{Metfaces (1.3K)~\cite{Tero-2020-stylegan3}}                            & \multicolumn{6}{c}{None}                                                                                                                                                              \\ \bottomrule[1.3pt]
\end{tabular}
}
\label{stylegan_detailed}
\end{table*}

We detailed the collection process of our datasets in Section 2.2 of the main paper, now the following will provide more detailed configuration information for each dataset.

We use the default Github repository code of each model to generate our datasets.
Detailed information about the training and validation datasets are shown in Tab.~\ref{model_dataset_detail}.
We further provide the captions and resolutions used in each specific dataset.
For diffusion generation, we use the fixed seed and cfg-scale to generate our datasets. We also use different sampling methods and steps for generation. The detailed information about sampling methods and steps for different diffusion models can be found in Tab.~\ref{diffusion_detailed}.
For StyleGAN3 generation, we use 2 models (stylegan3-r-ffhqu-1024x1024, stylegan3-t-ffhqu-1024x1024) to generate 70K face images for training to match the number of FFHQ and 10K face images for testing. We use 2 models (stylegan3-r-afhqv2-512x512, stylegan3-t-afhqv2-512x512) to generate 16K animal faces to match the number of AFHQ-v2 and 10K animal faces for testing. We use 2 models (stylegan3-r-metfaces-1024x1024, stylegan3-t-metfaces-1024x1024) to generate 1.3K art human faces for training to match the number of MetFaces Dataset and 10K art human faces for testing. The detailed information about our StyleGAN3 generation can be found in Tab.~\ref{stylegan_detailed}.




\subsection{Data Content Component Analysis of Training and Validation Dataset}

We will analyze the composition of the training and validation dataset in the following two parts and discuss the issue of data imbalance.
We also provide a detailed table showing the composition and proportion of different datasets, as shown in Tab.~\ref{dataset_content}.


\noindent{\textbf{Training Dataset.}}

\noindent{\textbf{$\bullet$}}
\textbf{Fake2M} is composed of 1M fake images generated by the first 1M caption in CC3M using SD-V1.5Real-dpms-25~\cite{Robin-2022-sd}, 1M fake images generated by the first 1M caption in CC3M using IF-V1.0-dpms++-25~\cite{if} and 87K fake images generated using StyleGAN3~\cite{Tero-2020-stylegan3}, as shown in Tab.~\ref{model_dataset_detail} and Tab~\ref{diffusion_detailed}.

In Fake2M, the number of face data is only 82K, accounting for \%4 of the total data 2M, as shown in Tab.~\ref{dataset_content}.
There is no content imbalanced problem in Fake2M.

\noindent{\textbf{$\bullet$}}
\textbf{Training Dataset Setting A} is composed of 1M fake images generated by the first 1M caption in CC3M using SD-V1.5Real-dpms-25 in Fake2M and the first 1M real images in CC3M.

In Training Dataset Setting A, most of the content is general content.
There is no content imbalanced problem in Training Dataset Setting A.

\noindent{\textbf{$\bullet$}}
\textbf{Training Dataset Setting B} is composed of 1M fake images generated by the first 1M caption in CC3M using IF-V1.0-dpms++-25 in Fake2M and the first 1M real images in CC3M.

In Training Dataset Setting B, most of the content is general content.
There is no content imbalanced problem in Training Dataset Setting B.

\noindent{\textbf{$\bullet$}}
\textbf{Training Dataset Setting C} is composed of 87K fake images generated by StyleGAN3 in Fake2M (the detailed content in this dataset can be found in Tab.~\ref{stylegan_detailed}) and the first 1M real images in CC3M.

In training dataset setting C, most of the content is face.
There is content imbalanced problem in training dataset setting C.
This inclusion was intentional, aiming to specifically investigate the performance implications of face fake images produced by StyleGAN3.

\noindent{\textbf{$\bullet$}}
\textbf{Training Dataset Setting D} is composed of 460K fake images generated by the first 460K caption in CC3M using IF-V1.0-dpms++-25 in Fake2M, 460K fake images generated by the first 460K caption in CC3M using SD-V1.5Real-dpms-25 in Fake2M, 87K fake images generated by StyleGAN3, the first 1M real images in CC3M and 87K real images in StyleGAN3 training dataset.

In training dataset setting D, most of the content is general content.
The number of fake face data is 82K and real face data is also 82K, accounting for \%8 of the total data 2M.
There is no content imbalanced problem in training dataset setting D.

\noindent{\textbf{Validation Dataset (MPBench).}}
In MPBench, most of the content is general content.
The number of fake face data is 60K, accounting for \%15.3 of the total data 391.5K.
The number of real face data is 24K, accounting for \%6.1 of the total data 391.5K.
There is no content imbalanced problem in training dataset setting D.

From the perspective of the ratio between real and fake images, we observe that the proportion of real and fake images is essentially the same across the four dataset settings and MPBench, as shown in Tab.~\ref{dataset_content}. Therefore, there is no imbalance issue between the number of fake and real images.

\begin{table*}[h]
\vspace{-0.2cm}
\caption{
\textbf{Data Content Component Analysis.}
"Content" means the type of the content in each dataset ("Face" means that the content in this dataset is mostly faces, such as FFHQ~\cite{Tero-2019-stylegan}. "Object" means that the content in this dataset is mainly composed of a limited number of objects, such as ImageNet~\cite{jia-2009-imagenet}. "General" means that the content in this dataset is general, not limited to some objects, faces or art, such as CC3M~\cite{Piyush-2018-cc3m}). "Each Dataset / Total Number (\%)" means the number of images in this dataset and the percentage it contributes to the entire dataset setting setting. "Fake / Total Number (\%)" means the number of fake images in the whole dataset setting and the percentage it contributes to the entire dataset setting. "Real / Total Number (\%)" means the number of real images in the whole dataset setting and the percentage it contributes to the entire dataset setting.
}
\centering
\renewcommand\arraystretch{1.4}
\resizebox{1\linewidth}{!}{

\begin{tabular}{l|llll|llll}
\toprule[1.3pt]
\multirow{2}{*}{\textbf{Dataset}}                       & \multicolumn{4}{c|}{\textbf{Fake}}                                                               & \multicolumn{4}{c}{\textbf{Real}}                                                          \\ \cline{2-9} 
                                               & \textbf{Name}                & \textbf{Content} & \textbf{Each Dataset / Total Number (\%)} & \textbf{Fake / Total Number (\%)}         & \textbf{Name}            & \textbf{Content} & \textbf{Each Dataset / Total Number (\%)} & \textbf{Real / Total Number (\%)}       \\ \midrule[1.3pt]
\multirow{3}{*}{Fake2M Dataset}                & SD-V1.5Real-dpms-25 & General & 1M (47.9\%)         & \multirow{3}{*}{2.08M (100\%)}    &                 &         &                     &                                 \\
                                               & IF-V1.0-dpms++-25   & General & 1M (47.9\%)         &                                   &                 &         &                     &                                 \\
                                               & StyleGAN3           & Face    & 87K (4.2\%)         &                                   &                 &         &                     &                                 \\ \hline
Training Dataest Setting A                     & SD-V1.5Real-dpms-25 & General & 1M (50\%)           & 1M (50\%)                         & CC3M-Train      & General & 1M (50\%)           & 1M (50\%)                       \\ \hline
Training Dataest Setting B                     & IF-V1.0-dpms++-25   & General & 1M (50\%)           & 1M (50\%)                         & CC3M-Train      & General & 1M (50\%)           & 1M (50\%)                       \\ \hline
Training Dataest Setting C                     & StyleGAN3           & Face    & 87K (50\%)          & 87K (50\%)                        & CC3M-Train      & General & 87K (50\%)          & 87K (50\%)                      \\ \hline
\multirow{3}{*}{Training Dataest Setting D}    & SD-V1.5Real-dpms-25 & General & 460K (21.2\%)       & \multirow{3}{*}{1.08M (50\%)}     & CC3M-Train      & General & 1M (46\%)           & \multirow{3}{*}{1.08M (50\%)}   \\
                                               & IF-V1.0-dpms++-25   & General & 460K (21.2\%)       &                                   & StyleGAN3-Train & Face    & 87K (4\%)           &                                 \\
                                               & StyleGAN3           & Face    & 87K (4\%)           &                                   &                 &         &                     &                                 \\ \midrule[1.3pt]
\multirow{11}{*}{Validation Dataset (MPBench)} & SD-V2.1-dpm-25      & General & 15K (3.8\%)         & \multirow{11}{*}{252.5K (64.5\%)} & ImageNet-Test   & Object  & 100K (25.5\%)       & \multirow{11}{*}{139K (35.5\%)} \\
                                               & SD-V1.5-dpm-25      & General & 15K (3.8\%)         &                                   & CelebA-HQ-Train & Face    & 24K (6.1\%)         &                                 \\
                                               & SD-V1.5Real-dpm-25  & General & 15K (3.8\%)         &                                   & CC3M-Val        & General & 15K (3.8\%)         &                                 \\
                                               & IF-V1.0-dpm++-10    & General & 15K (3.8\%)         &                                   &                 &         &                     &                                 \\
                                               & IF-V1.0-dpm++-25    & General & 15K (3.8\%)         &                                   &                 &         &                     &                                 \\
                                               & IF-V1.0-dpm++-50    & General & 15K (3.8\%)         &                                   &                 &         &                     &                                 \\
                                               & IF-V1.0-ddim-50     & General & 15K (3.8\%)         &                                   &                 &         &                     &                                 \\
                                               & IF-V1.0-ddpm-50     & General & 15K (3.8\%)         &                                   &                 &         &                     &                                 \\
                                               & Cogview2            & General & 22K (5.6\%)         &                                   &                 &         &                     &                                 \\
                                               & Midjourney          & General & 5.5K (1.4\%)        &                                   &                 &         &                     &                                 \\
                                               & StyleGAN3           & Face    & 60K (15.3\%)        &                                   &                 &         &                     &                                 \\ \bottomrule[1.3pt]
\end{tabular}
}
\label{dataset_content}
\end{table*}

\subsection{Quality Analysis}
\begin{figure*}[!h]
    \centering    \includegraphics[width=1\linewidth]{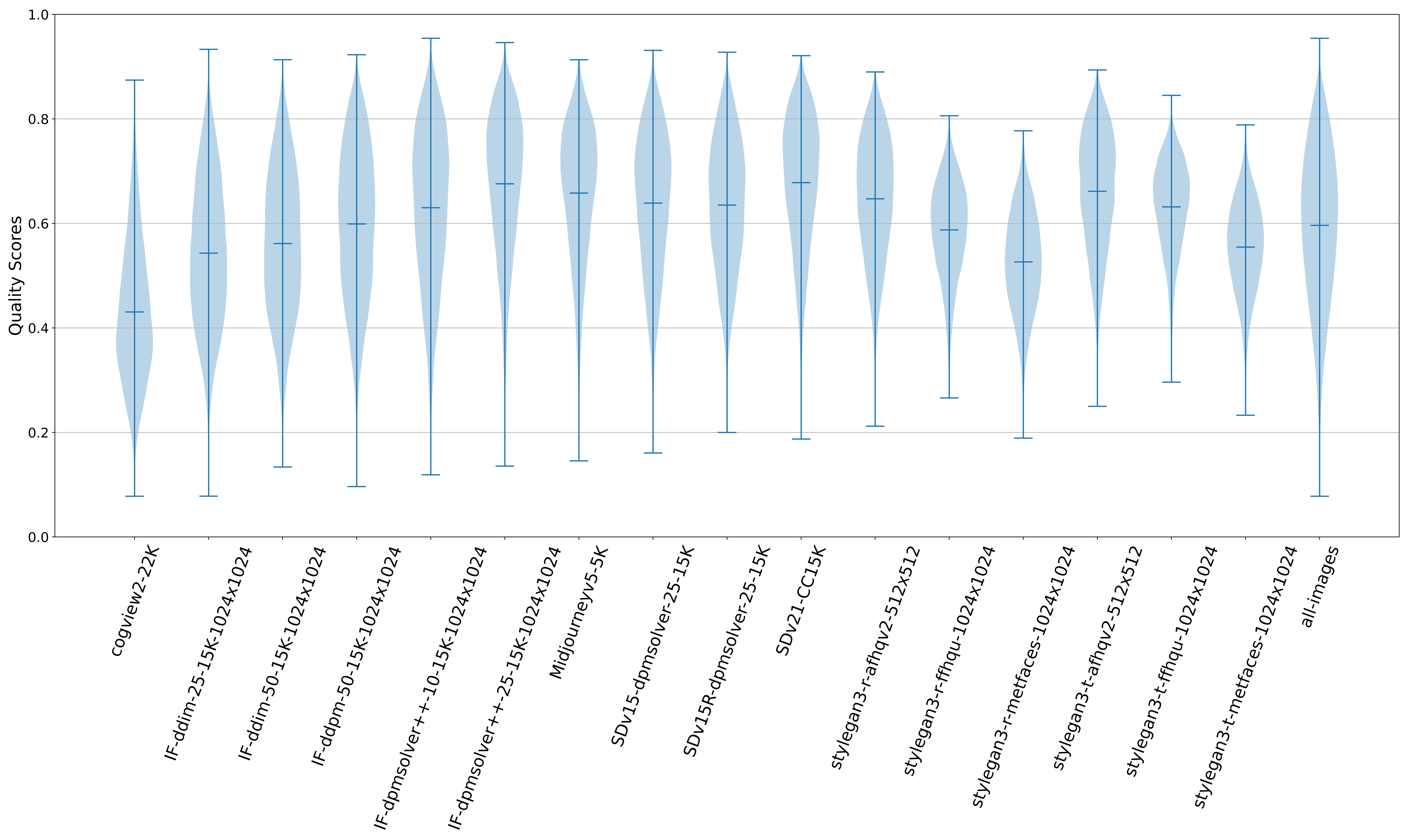}
    \caption{
        \textbf{Quality score distribution of the dataset.} "all-images" means the quality score distribution of all the images in the dataset.
    }
    \label{quality_score_distribution}
\end{figure*}

\begin{figure*}[!h]
    \centering    \includegraphics[width=1\linewidth]{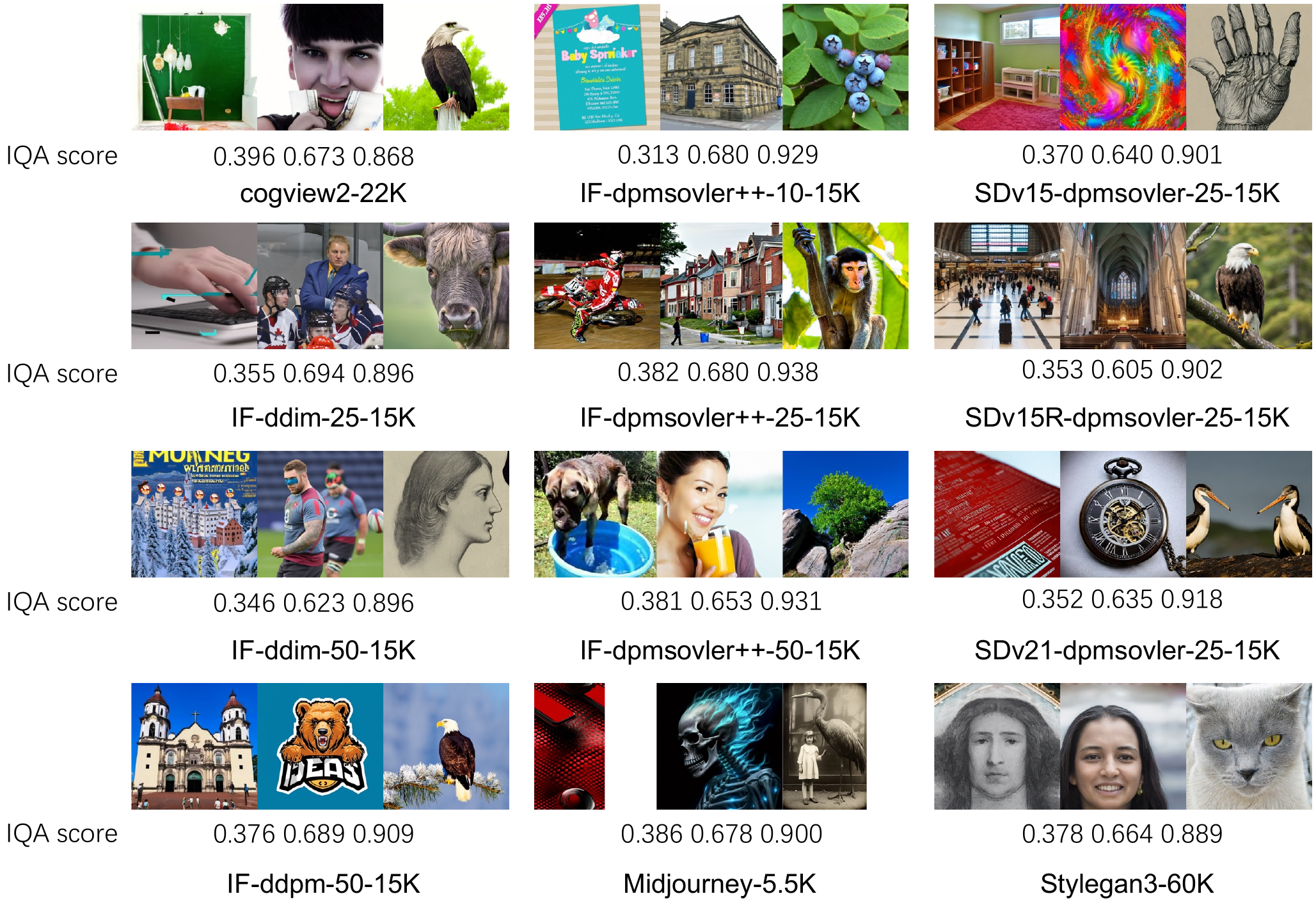}
    \caption{
        \textbf{Image visualization with image quality score.}
    }
    \label{quality_visulization}
\end{figure*}

We conducted further analysis of our dataset quality score distributions, as shown in Fig.~\ref{quality_score_distribution} and Tab.~\ref{quality_distribution}. We observed that the majority of our sub dataset have an average score above 0.6 (with Midjourneyv5-5K having an average score of 0.66) and the average score of all images in the dataset is 0.6. These demonstrate that our dataset is a high-quality dataset with a large amount of high-quality images. Only a few datasets (cogview2-22K, IF-ddim-25-15K-1024x1024, IF-ddim-50-15K-1024x1024, stylegan3-r-ffhqu-1024x1024, and stylegan3-r-metfaces-1024x1024) have an average score below 0.6. The distribution of quality scores across the entire dataset demonstrates a balanced mixture of high-quality and low-quality images, as shown in the "all-images" violin plot of Fig.~\ref{quality_score_distribution}.  This aligns with our original intention: a fake image detection dataset should encompass both high-quality and low-quality image data. In order to better showcase our dataset, we provided more visualizations about the high quality, mid quality and low quality images in our dataset, as shown in Fig.~\ref{quality_visulization}.

\begin{table*}[!h]
\caption{\textbf{Quality score distribution statistical information of the dataset.} "all-images" means the quality score distribution of all the images in the dataset. "Mean Score" means the average score of the quality score in the sub dataset.
"Min Score" means the minimum score of the quality score in the sub dataset.
"Max Score" means the maximum score of the quality score in the sub dataset.
}
\centering
\renewcommand\arraystretch{1.4}
\resizebox{0.83\linewidth}{!}{
\begin{tabular}{c|ccc}
\toprule[1.2pt]
\textbf{Sub Dataset}                     & \textbf{Mean Score} & \textbf{Min Score} & \textbf{Max Score} \\ \hline
cogview2-22K                    & 0.43       & 0.08      & 0.87      \\
IF-ddim-25-15K-1024x1024        & 0.54       & 0.08      & 0.93      \\
IF-ddim-50-15K-1024x1024        & 0.56       & 0.13      & 0.91      \\
IF-ddpm-50-15K-1024x1024        & 0.60       & 0.10      & 0.92      \\
IF-dpmsolver++-10-15K-1024x1024 & 0.63       & 0.12      & \textbf{0.95}      \\
IF-dpmsolver++-25-15K-1024x1024 & 0.68       & 0.14      & \textbf{0.95}      \\
Midjourneyv5-5K                 & 0.66       & 0.15      & 0.91      \\
SDv15-dpmsolver-25-15K          & 0.64       & 0.16      & 0.93      \\
SDv15R-dpmsolver-25-15K         & 0.64       & 0.20      & 0.93      \\
SDv21-CC15K                     & \textbf{0.68}       & 0.19      & 0.92      \\
stylegan3-r-afhqv2-512x512      & 0.65       & 0.21      & 0.89      \\
stylegan3-r-ffhqu-1024x1024     & 0.59       & 0.27      & 0.81      \\
stylegan3-r-metfaces-1024x1024  & 0.53       & 0.19      & 0.78      \\
stylegan3-t-afhqv2-512x512      & 0.66       & 0.25      & 0.89      \\
stylegan3-t-ffhqu-1024x1024     & 0.63       & \textbf{0.30}      & 0.85      \\
stylegan3-t-metfaces-1024x1024  & 0.55       & 0.23      & 0.79      \\ \hline
all-images                      & 0.60       & 0.08      & \textbf{0.95}      \\ \bottomrule[1.2pt]
\end{tabular}
}
\label{quality_distribution}
\end{table*}

\newpage

\section{HPBench}

\subsection{Procedures for HPBench}
\begin{figure*}[!h]
    \centering    \includegraphics[width=1\linewidth]{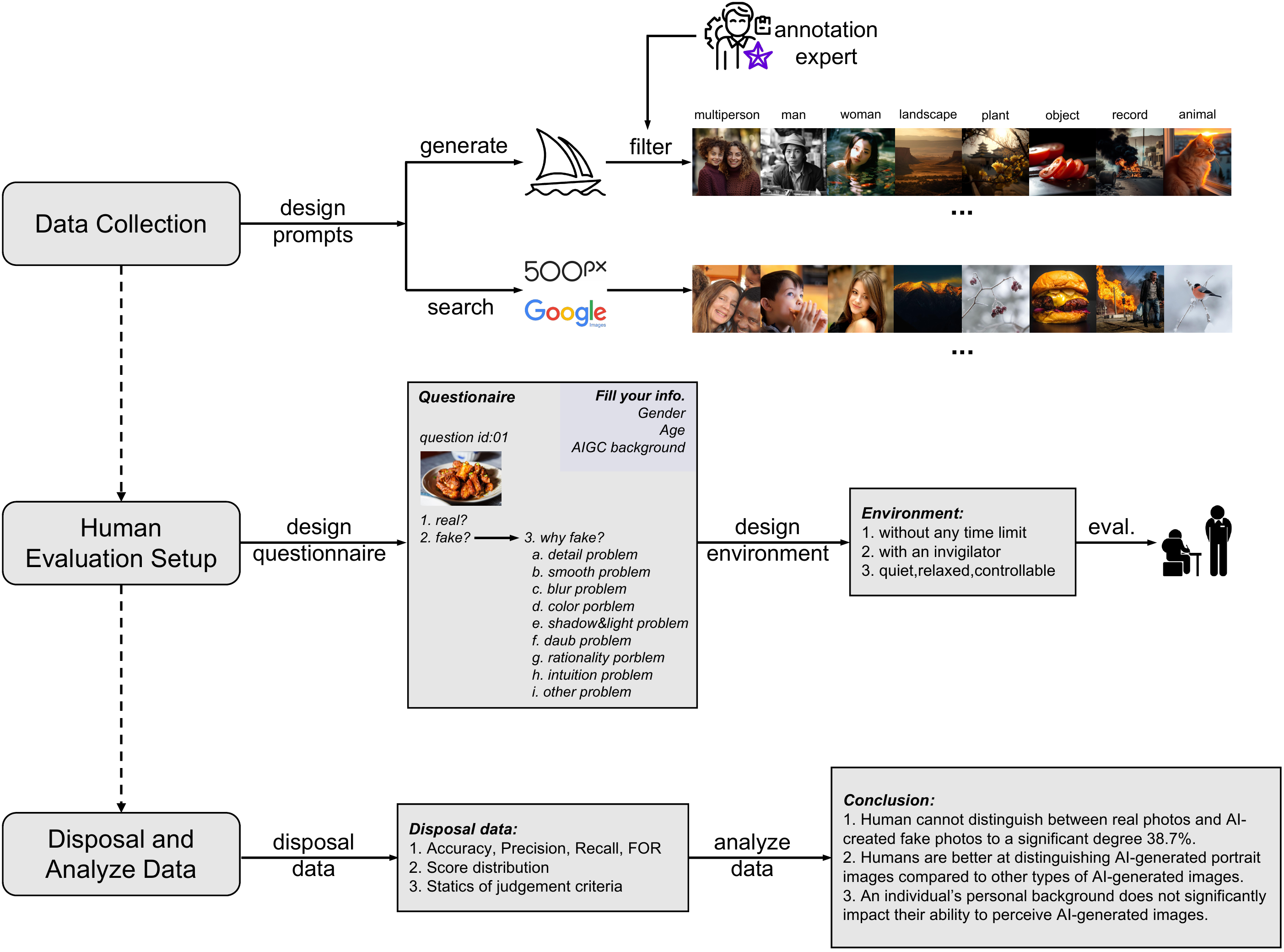}
    \vspace{-0.1cm}
    \caption{
        \textbf{Procedures for HPBench.} 
    }
    \vspace{-0.3cm}
    \label{pipeline}
\end{figure*}

Fig.~\ref{pipeline} shows that our HPBench could be divided into three parts. 
In the first part, we collect realistic AI-generated images and real images across eight categories using the expertise of an annotator to filter out low-quality AI-generated images. 
In the second part, we recruit a total of 50 volunteers for human evaluation.
For each volunteer, he/she should complete a 100-question questionnaire in our prepared environment.
In the third part, We disposal and analyze human evaluation data to draw conclusions.

\subsection{Detail of Human Evaluation}
We recruit a total of 50 participants to participate in our human evaluation instead of using crowdsourcing. 
In order to ensure comprehensiveness, fairness, and quality of the evaluation, we make efforts to ensure the diversity of the participants. 
Each participant is asked to complete a questionnaire consisting of 100 questions without any time limit. 
The questionnaires are completed in the presence of a project team member to guide the participant and ensure the quality of the human evaluation results.
It is worth noting that we did not inform the participant about the ratio of real photos to AI-generated images in the questionnaires.

Specifically, each question in the questionnaire provides the participant with an image, and the participant is asked to determine whether the image is generated by AI or not. If the participant thinks that the image is generated by AI, he/she will be required to choose one or more reasons from the eight predefined judgment criteria options or provide their own judgment criteria. The eight options are explained below:

\noindent{\textbf{$\bullet$}} Detail: AI-generated images may lack fine details, such as wrinkles in clothing or hair details.

\noindent{\textbf{$\bullet$}} Smooth: AI-generated images may appear smoother or more uniform than real photos, such as smooth skin or unrealistic facial expressions.

\noindent{\textbf{$\bullet$}} Blur: AI-generated images may be blurry, such as blurry or unclear edges.

\noindent{\textbf{$\bullet$}} Color: AI-generated images may have unrealistic or inconsistent colors, such as colors that are too bright, too dark, or like the color of animation.

\noindent{\textbf{$\bullet$}} Shadow \& Light: AI-generated images may have unrealistic or inconsistent shadows and lighting, such as shadows or lightning that violate physics.

\noindent{\textbf{$\bullet$}} Daub: AI-generated images may contain rough, uneven, or poorly applied colors or textures.

\noindent{\textbf{$\bullet$}} Rationality: AI-generated images may contain irrational/illogical/contradictory contents.

\noindent{\textbf{$\bullet$}} \textcolor{MyDarkgray}{Intuition}: people may judge whether a photo is AI-generated or not by intuition and cannot describe the exact reasons.

It is important to point out that each questionnaire consists of 50 real images and 50 AI-generated images, all of which are randomly sampled from the real image database containing 244 images and the AI-generated image database containing 151 images.

\subsection{Reason of High-quality Fifty-participant Human Evaluation}
Inspired by Robert \textit{et al.}~\cite{Robert-2021-humanvsmachine}, we aim to collect high-quality human evaluation data instead of noisy crowdsourcing data to ensure the high quality of the results. Our human evaluation, with a limited number of participants in a controlled environment, and conducting multiple experiments, is commonly referred to as the ``small-N design''. As the article "Small is beautiful: In defense of the small-N design" \cite{Smith-2018-small} suggests, the "small-N design" is the core of high-quality psychophysics. Crowdsourcing involves many participants in an uncontrollable, noisy setting, with each performing fewer trials. In contrast, we strive to ensure that each participant, with diverse backgrounds, takes the full 100-question survey in a quiet, relaxed, controllable, and monitored environment. Those factors contribute to high data quality.

\subsection{Crowd Sourcing Human Evaluation}
We collect 1085 crowd-sourced human evaluation questionnaires to make the entire benchmark more comprehensive. 
We utilize the same experimental setup as the "High-quality Fifty-participant Human Evaluation" before.
As the questionnaires are obtained through crowd-sourcing in an uncontrollable and noisy setting, we do not ask the participants to provide justification for each decision.
We collect the accuracy of all the questionnaires, and the accuracy is only 49.9\%.
This highlights that in a fast-paced, noisy, and uncontrolled environment, people are completely unable to distinguish high-quality AI-generated images.

\definecolor{red}{RGB}{255,64,64}
\begin{figure*}[!t]
    \centering
    \includegraphics[width=0.9\linewidth]{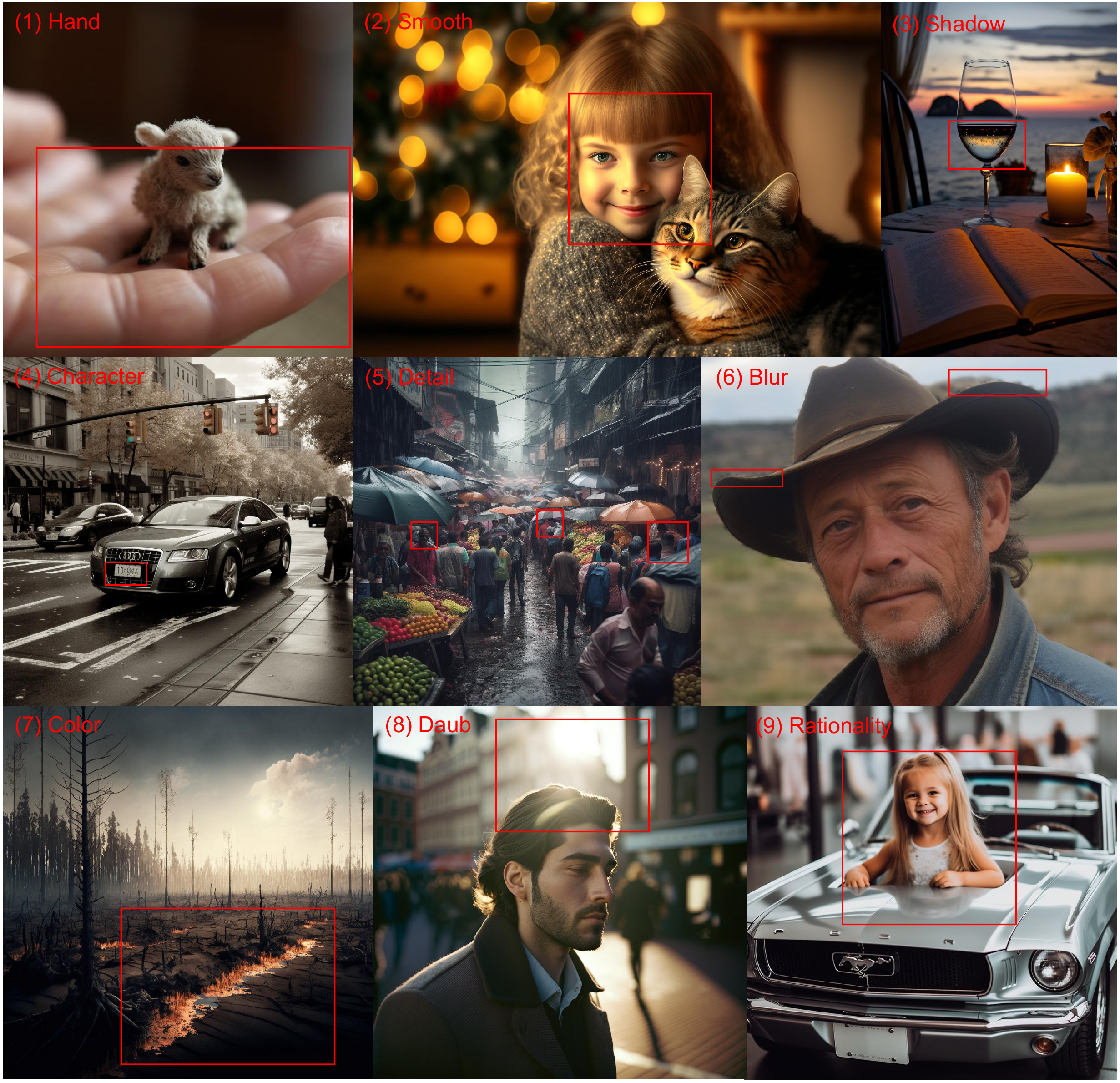}
    \caption{
        \textbf{Nine shortcomings
of current AIGC.} We highlight the obvious defects of AIGC with \textcolor{red}{red boxes}.
    }
    \label{AIGC defects}
\end{figure*}

\subsection{Metrics of HPBench}
We employ four commonly used evaluation metrics to analyze our results and highlight their respective meanings in the context of our problem. We define positive samples as AI-generated images and negative samples as real images for our problem, and then calculate Accuracy, Precision, Recall, and False Omission Rate (FOR) in the context of our problem.

\textbf{Accuracy}
is a statistical measure used to evaluate how well a binary classification test correctly identifies or excludes a condition. In our study, accuracy represents the average precision of humans in distinguishing AI-generated images from real images.

\textbf{Precision}
is the percentage of predicted positive cases that are actually positive. In our study, high precision represents the proportion of AI-generated images out of the total number of images that are predicted as AI-generated ones. 

\textbf{Recall}
is the percentage of true positive cases that are actually predicted as positive. In our study, recall represents the proportion of AI-generated images that are correctly identified as such out of the total number of AI-generated images.

\textbf{FOR}
is the percentage of false negatives out of all negative cases. In our problem, FOR represents the proportion of real images misidentified as AI-generated images out of the total number of images.


\subsection{Analysis of the AIGC defects}

Based on the user data we collected above, we summarize and show nine shortcomings of the current AIGC, as shown in Fig.~\ref{AIGC defects}:
(1) "Hand problem" refers to situations where fingers overlap or have unreasonable shapes (multi fingers), resulting in images that are not realistic.
(2) "Smoothing problem" refers to situations where AI-generated images have overly smooth skin, resulting in unrealistic facial expressions and features.
(3) "Shadow\&Light problem" refers to situations where the position and shape of light sources and shadows in AI-generated images are unreasonable, resulting in unnatural lighting effects.
(4) "Character problem" refers to situations where incorrect signs and texts appear in AI-generated images, which do not match reality.
(5) "Detail problem" refers to situations where some details in AI-generated images are not realistic or unreasonable, such as wrinkles in clothing or hair details.
(6) "Blur problem" refers to situations where AI-generated images are blurry or unclear, resulting in obvious artifacts.
(7) "Color problem" refers to situations where the colors in AI-generated images are not realistic or coordinated, such as colors that are too bright, too dark, or like the color of animation.
(8) "Daub problem" refers to situations where AI-generated images have been excessively daubed, resulting in lost details or unrealistic images.
(9) "Rationality problem" between objects refers to situations where the relationship between objects in AI-generated images is not reasonable, such as incorrect size proportions or unreasonable positions.

\begin{wrapfigure}{!R}{0.32\textwidth}
    \centering
    \includegraphics[width=0.99\linewidth]{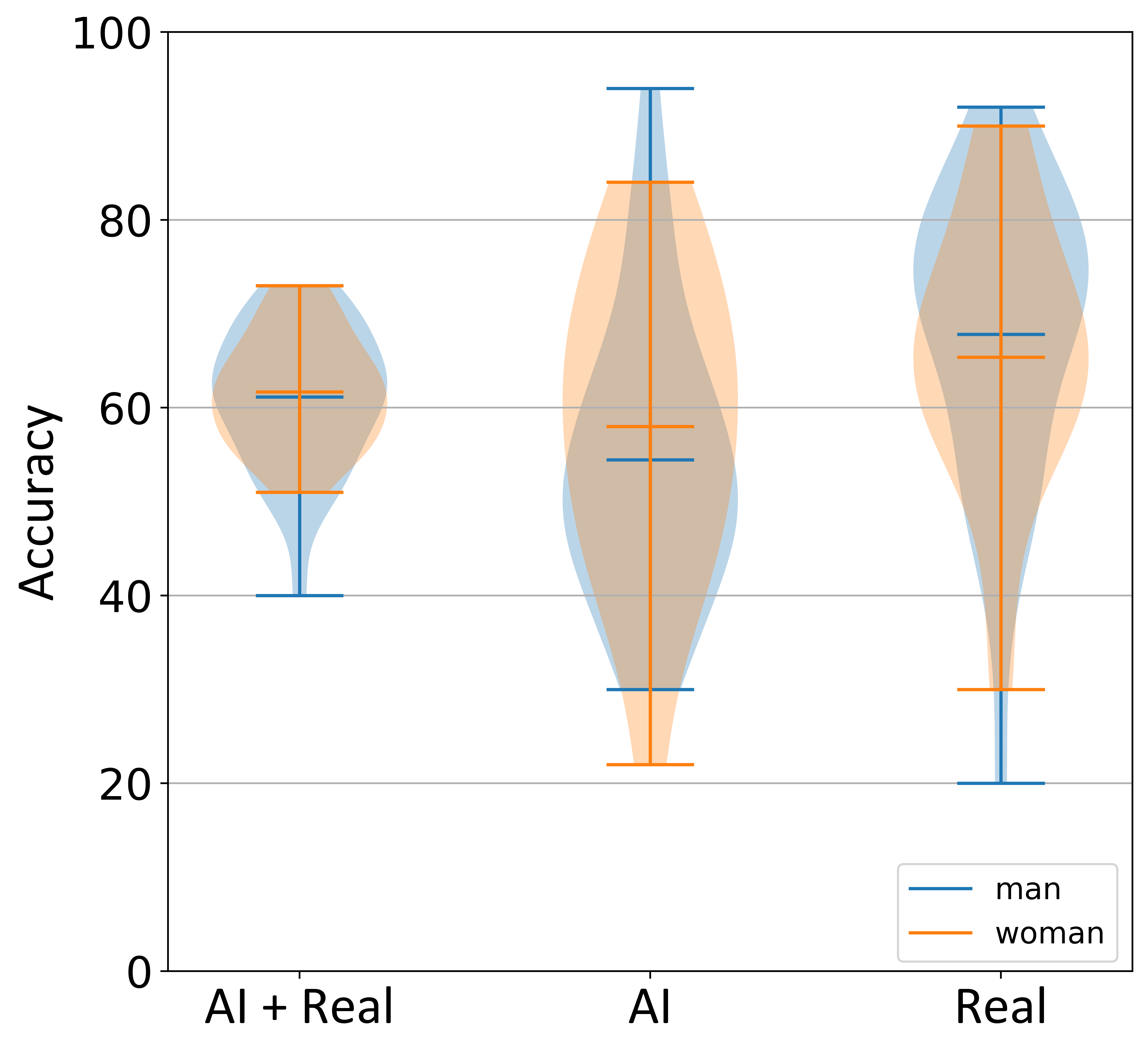}
    \caption{
        \textbf{
        Human evaluation score distributions calculated by data from men and women.
        }
    }
    \label{all categories with man and woman}
\end{wrapfigure}

\subsection{Detailed Score Distribution}
\begin{figure*}[!h]
    \centering
    \includegraphics[width=1\linewidth]{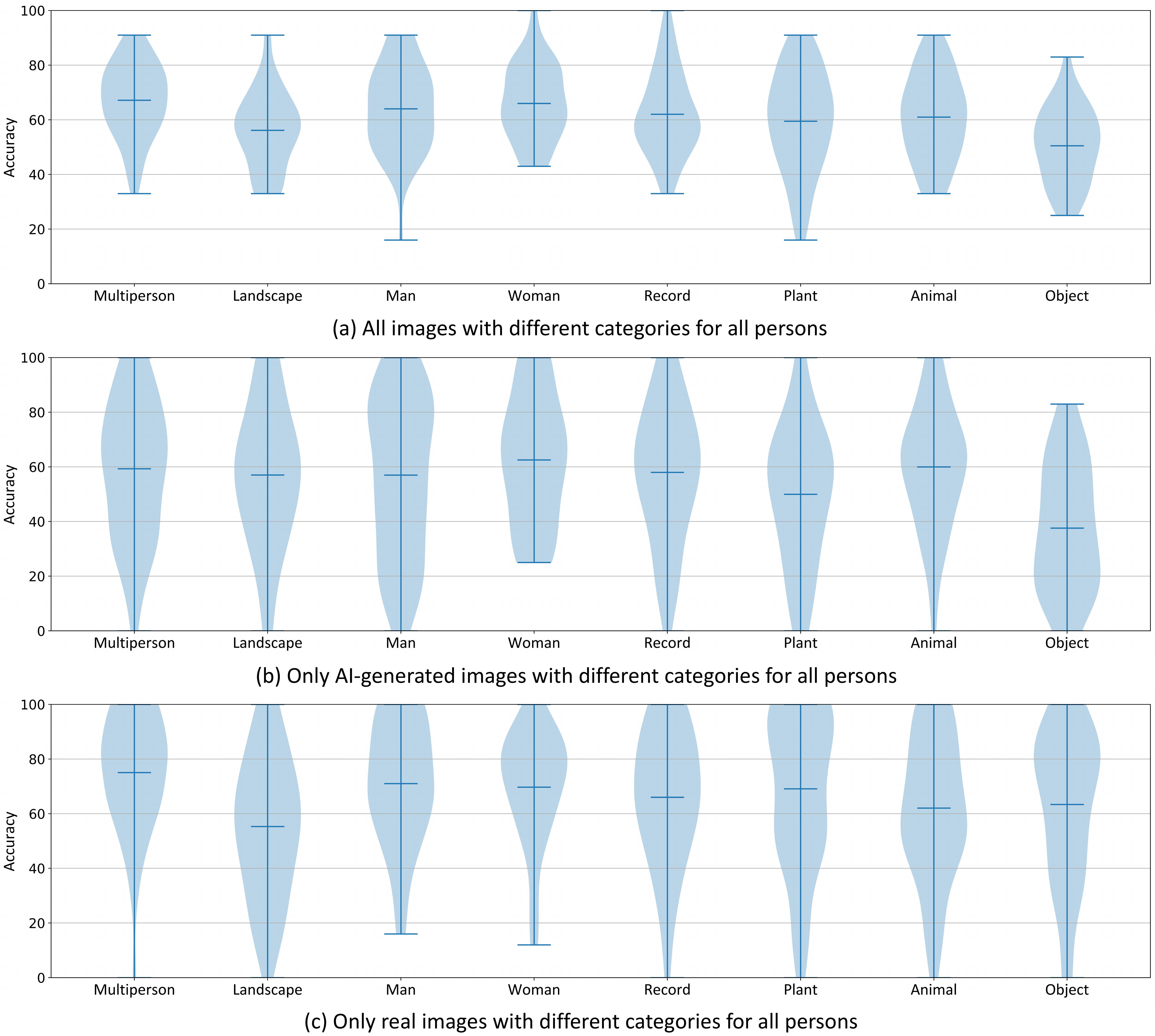}
    \caption{
        \textbf{Score distributions for all volunteers with different categories.}
    }
    \label{distribution of different categories for all persons}
\end{figure*}

\paragraph{Detailed score distribution of different categories for all volunteers.}
As shown in Fig.~\ref{distribution of different categories for all persons}, we visualize the detailed score distribution of different categories for all volunteers: (a) the detailed score distribution of different categories for all volunteers and all tested images (b) the detailed score distribution of different categories for all volunteers and only AI-generated images (c) the detailed score distribution of different categories for all volunteers and only real images.
\paragraph{Detailed score distribution of different categories for men and women.}
As shown in Fig.~\ref{all categories with man and woman}, we visualize the score distribution of all images for man and woman.
We find that the average scores of men and women are almost the same with a relatively accuracy rate of 61\%.

\begin{figure*}[!h]
    \centering
    \includegraphics[width=1\linewidth]{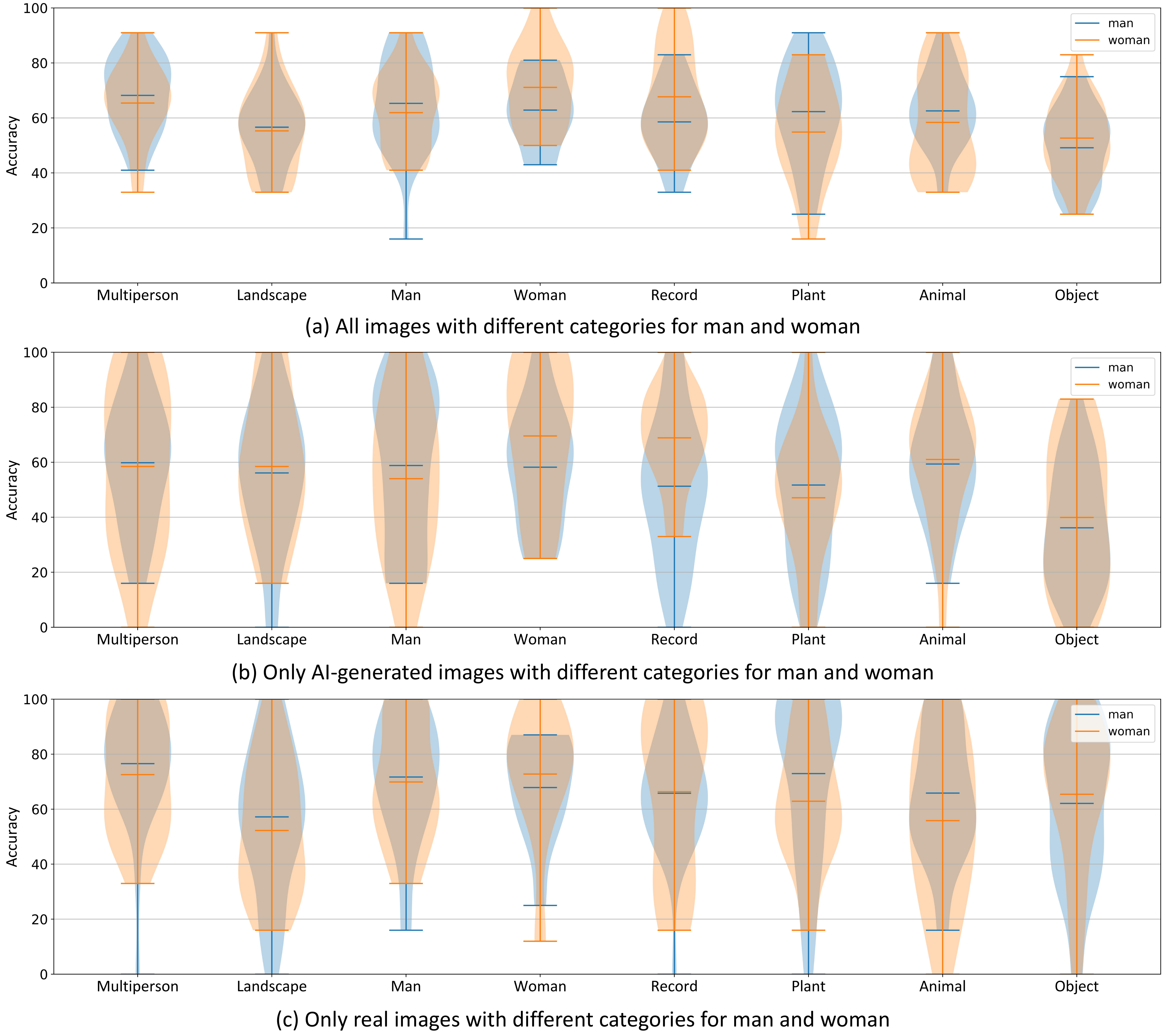}
    \caption{
        \textbf{Score distributions for men and women with different categories.}
    }
    \label{distribution of different categories for man and woman}
\end{figure*}

\begin{figure*}[!h]
    \centering
    \includegraphics[width=1\linewidth]{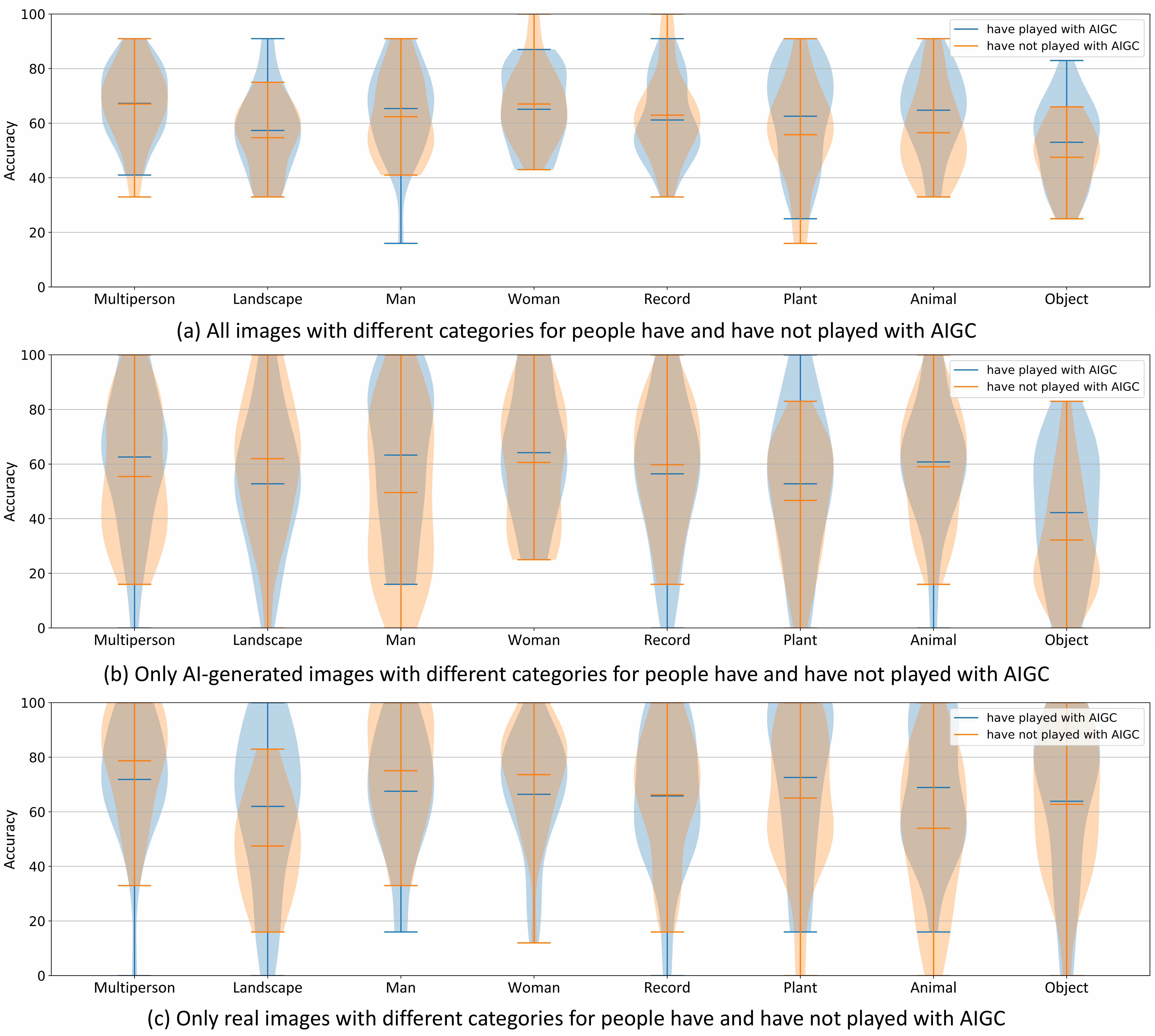}
    \caption{
        \textbf{Score distributions for volunteers with and without AIGC background.}
    }
    \label{distribution of different categories for AIGC}
\end{figure*}

We also visualize the detailed score distribution of different categories for man and woman in Fig.~\ref{distribution of different categories for man and woman}: (a) the detailed score distribution of different categories for man and woman and all tested images (b) the detailed score distribution of different categories for man and woman and only AI-generated images (c) the detailed score distribution of different categories for man and woman and only real images. A interesting finding is that: Apart from humans having a high recognition rate for human portraits, men have a higher recognition rate for the category \textit{Man} than women, and women have a higher recognition rate for the category \textit{Woman} than men. We speculate that people may have a higher recognition accuracy for more familiar objects.
\paragraph{Detailed score distribution of different categories for volunteers with and without AIGC background.}
As shown in Fig.~\ref{distribution of different categories for AIGC}, we visualize the detailed score distribution of different categories for volunteers with and without AIGC background: (a) the detailed score distribution of different categories for volunteers with and without AIGC background and all tested images (b) the detailed score distribution of different categories for volunteers with and without AIGC background and only AI-generated images (c) the detailed score distribution of different categories for volunteers with and without AIGC backgrounds and only real images.

\section{MPBench}
\begin{table}[!h]
\renewcommand\arraystretch{1.3}
\setlength\tabcolsep{2pt}
\caption{
\textbf{Quantitative comparison of another five models under four training dataset settings with fourteen validation datasets.}
"Diff" refers to diffusion model, "AR" refers to autoregressive model and "Unk." refers to unknown model.
{\color[HTML]{CB0000} \textbf{Real (R)}} denotes the dataset consisting entirely of real images.
{\color[HTML]{036400} \textbf{Fake (F)}} denotes the dataset consisting entirely of fake images.
{\color[HTML]{AAEEFF} Blue cell} denotes the deepfake methods.
}
\resizebox{1\linewidth}{!}{
\begin{tabular}{@{}l|c|cccc|cccccccccccc|c@{}}
\toprule[1.5pt]
&       & \rotatebox[origin=lb]{90} {\smash{\small ImageNet-Test}} & \rotatebox[origin=lb]{90} {\smash{\small CelebA-HQ-Train}} & \rotatebox[origin=lb]{90} {\smash{\small CC3M-Val}} & \cellcolor[HTML]{EFEFEF}\rotatebox[origin=lb]{90} {\smash{\small Average Acc.}}  & \rotatebox[origin=lb]{90} {\smash{\small SD-V2.1-dpm-25 }} & \rotatebox[origin=lb]{90} {\smash{\small SD-V1.5-dpm-25 }} & \rotatebox[origin=lb]{90} {\smash{\small SD-V1.5Real-dpm-25 }} & \rotatebox[origin=lb]{90} {\smash{\small IF-V1.0-dpm++-10 }} & \rotatebox[origin=lb]{90} {\smash{\small IF-V1.0-dpm++-25 }} & \rotatebox[origin=lb]{90} {\smash{\small IF-V1.0-dpm++-50 }} & \rotatebox[origin=lb]{90} {\smash{\small IF-V1.0-ddim-50 }} & \rotatebox[origin=lb]{90} {\smash{\small IF-V1.0-ddpm-50 }} & \rotatebox[origin=lb]{90} {\smash{\small Cogview2 }} & \rotatebox[origin=lb]{90} {\smash{\small Midjourney }} & \rotatebox[origin=lb]{90} {\smash{\small StyleGAN3 }} & \cellcolor[HTML]{EFEFEF} \rotatebox[origin=lb]{90} {\smash{\small Average Acc. }}& \cellcolor[HTML]{FFF8F8} \rotatebox[origin=lb]{90} {\smash{\small Total Average Acc. }} \\ 
\multicolumn{1}{l|}{}                        & \multicolumn{1}{c|}{}                               & \multicolumn{4}{c|}{{\color[HTML]{CB0000} \textbf{Real}}}                                                              & \multicolumn{12}{c|}{{\color[HTML]{036400} \textbf{Fake}}}                                                                                                                                                                                                                                                      &  {\color[HTML]{CB0000} \textbf{R}}+{\color[HTML]{036400} \textbf{F}}         \\
\multicolumn{1}{l|}{\multirow{1}{*}{\textbf{Model}}} & \multicolumn{1}{c|}{\multirow{1}{*}{\textbf{Training Dataset}}}        & -                    & -                    & -                    & \multicolumn{1}{c|}{-}    & Diff.                & Diff.                & Diff.                & Diff.                & Diff.                & Diff.                & Diff.                & Diff.                & AR                   & Unk.                 & GAN                  & \multicolumn{1}{c|}{-}     & -         \\ \midrule[1.5pt]
\multicolumn{1}{l|}{Swin-S(B+J 0.1)}     & \multicolumn{1}{c|}{}                               & 92.4                 & 100.0                 & 99.9                 & \multicolumn{1}{c|}{\cellcolor[HTML]{EFEFEF}97.4} & 37.9                 & 71.7                 & 99.9                 & 94.4                 & 69.6                 & 69.4                 & 50.6                 & 76.9                 & 49.5                 & 6.8                 & 39.5                 & \multicolumn{1}{c|}{\cellcolor[HTML]{EFEFEF}60.5}  & \cellcolor[HTML]{FFF8F8}68.4      \\
\multicolumn{1}{l|}{Swin-S(B+J 0.5)}     & \multicolumn{1}{c|}{}                               & 95.1                 & 99.9                 & 99.9                 & \multicolumn{1}{c|}{\cellcolor[HTML]{EFEFEF}98.3} & 41.3                 & 60.2                  & 99.9                 & 89.7                 & 52.3                 & 52.2                 & 42.0                 & 67.6                 & 64.5                 & 9.5                 & 25.1                 & \multicolumn{1}{c|}{\cellcolor[HTML]{EFEFEF}54.9}  & \cellcolor[HTML]{FFF8F8}64.2      \\
\multicolumn{1}{l|}{DeiT-S(B+J 0.1)}       & \multicolumn{1}{c|}{}                               & 99.7                 & 99.9                 & 99.9                 & \multicolumn{1}{c|}{\cellcolor[HTML]{EFEFEF}99.8} & 37.8                 & 47.3                 & 99.9                 & 19.6                 & 2.6                 & 2.2                 & 5.2                 & 12.9                 & 8.5                 & 6.2                 & 2.1                  & \multicolumn{1}{c|}{\cellcolor[HTML]{EFEFEF}22.2}  & \cellcolor[HTML]{FFF8F8}38.8      \\
\multicolumn{1}{l|}{DeiT-S(B+J 0.5)}       & \multicolumn{1}{c|}{}                               & 99.4                 & 99.9                 & 99.8                 & \multicolumn{1}{c|}{\cellcolor[HTML]{EFEFEF}99.7} & 46.2                 & 51.5                 & 99.9                 & 21.5                 & 4.0                 & 3.1                 & 6.6                 & 13.9                 & 4.7                 & 8.6                 & 2.8                  & \multicolumn{1}{c|}{\cellcolor[HTML]{EFEFEF}23.8}  & \cellcolor[HTML]{FFF8F8}40.1      \\

\multicolumn{1}{l|}{ResNet50(Fourier)~\cite{Xu-2019-detectfake}}          & \multicolumn{1}{c|}{}         & 42.3                 & 7.2                 & 4.3                 & \multicolumn{1}{c|}{\cellcolor[HTML]{EFEFEF}17.9} & 95.9                 & 97.4                 & 97.3                 & 96.9                 & 96.0                 & 95.4                 & 98.6                 & 96.0                 & 95.6                 & 94.9                 & 97.0                 & \multicolumn{1}{c|}{\cellcolor[HTML]{EFEFEF}96.4}  & \cellcolor[HTML]{FFF8F8}79.6\\ 
\multicolumn{1}{l|}{Xception(Patch)~\cite{Lucy-2020-understand}}          & \multicolumn{1}{c|}{\multirow{-6}{*}{\makecell[c]{\textcolor{MyDarkgray}{Dataset Setting A:}\\ SD-V1.5Real-dpms-25 (1M)\\CC3M-Train (1M)}}}         & 57.1                 & 58.7                 & 55.7                 & \multicolumn{1}{c|}{\cellcolor[HTML]{EFEFEF}57.1} & 31.0                 & 34.9                 & 31.4                 & 36.5                 & 34.7                 & 36.9                 & 35.5                 & 37.6                 & 49.7                 & 39.5                 & 42.7                 & \multicolumn{1}{c|}{\cellcolor[HTML]{EFEFEF}37.3}  & \cellcolor[HTML]{FFF8F8}41.5\\ \midrule

\multicolumn{1}{l|}{Swin-S(B+J 0.1)}     & \multicolumn{1}{c|}{}                               & 88.1                 & 99.9                 & 99.9                 & \multicolumn{1}{c|}{\cellcolor[HTML]{EFEFEF}95.9} & 0.4                  & 3.0                 & 0.1                  & 99.6                 & 99.8                 & 99.7                 & 2.1                 & 22.0                 & 3.3                 & 0.1                  & 0.7                 & \multicolumn{1}{c|}{\cellcolor[HTML]{EFEFEF}30.0}  & \cellcolor[HTML]{FFF8F8}44.1      \\
\multicolumn{1}{l|}{Swin-S(B+J 0.5)}     & \multicolumn{1}{c|}{}                               & 81.5                 & 99.9                 & 99.9                 & \multicolumn{1}{c|}{\cellcolor[HTML]{EFEFEF}93.7} & 0.9                  & 9.9                 & 0.3                  & 99.6                 & 99.9                 & 99.8                 & 6.7                 & 42.6                 & 6.8                 & 0.1                  & 3.3                  & \multicolumn{1}{c|}{\cellcolor[HTML]{EFEFEF}33.6}  & \cellcolor[HTML]{FFF8F8}46.5      \\
\multicolumn{1}{l|}{DeiT-S(B+J 0.1)}       & \multicolumn{1}{c|}{}                               & 98.3                 & 99.7                 & 99.7                 & \multicolumn{1}{c|}{\cellcolor[HTML]{EFEFEF}99.2} & 8.9                 & 32.0                 & 9.5                 & 95.3                 & 99.2                 & 97.4                 & 53.6                 & 55.0                 & 25.4                 & 2.8                 & 2.5                  & \multicolumn{1}{c|}{\cellcolor[HTML]{EFEFEF}43.7}  & \cellcolor[HTML]{FFF8F8}55.6      \\
\multicolumn{1}{l|}{DeiT-S(B+J 0.5)}       & \multicolumn{1}{c|}{}                               & 97.8                 & 99.6                 & 99.4                 & \multicolumn{1}{c|}{\cellcolor[HTML]{EFEFEF}98.9} & 11.3                 & 36.6                 & 10.0                 & 95.2                 & 99.3                 & 97.9                 & 57.9                 & 61.9                 & 27.8                 & 3.7                 & 2.9                  & \multicolumn{1}{c|}{\cellcolor[HTML]{EFEFEF}45.8}  & \cellcolor[HTML]{FFF8F8}57.2      \\

\multicolumn{1}{l|}{ResNet50(Fourier)~\cite{Xu-2019-detectfake}}          & \multicolumn{1}{c|}{}           & 42.3                 & 45.7                 & 51.0                 & \multicolumn{1}{c|}{\cellcolor[HTML]{EFEFEF}46.3} & 60.7                 & 61.7                 & 73.0                 & 59.3                 & 72.2                 & 70.3                 & 29.5                 & 69.4                 & 40.8                 & 60.9                 & 70.5                 & \multicolumn{1}{c|}{\cellcolor[HTML]{EFEFEF}60.7}  & \cellcolor[HTML]{FFF8F8}57.6      \\ 
\multicolumn{1}{l|}{Xception(Patch)~\cite{Lucy-2020-understand}}          & \multicolumn{1}{c|}{\multirow{-5}{*}{\makecell[c]{\textcolor{MyDarkgray}{Dataset Setting B:}\\ IF-V1.0-dpms++-25 (1M)\\CC3M-Train (1M)}}}        & 54.5                 & 17.0                 & 29.0                 & \multicolumn{1}{c|}{\cellcolor[HTML]{EFEFEF}33.5} & 43.4                 & 44.1                 & 49.8                 & 23.7                 & 19.3                 & 20.1                 & 63.5                 & 35.8                 & 49.7                 & 57.0                 & 58.1                 & \multicolumn{1}{c|}{\cellcolor[HTML]{EFEFEF}42.2}  & \cellcolor[HTML]{FFF8F8}40.3\\ \midrule

\multicolumn{1}{l|}{Swin-S(B+J 0.1)}     & \multicolumn{1}{c|}{}                               & 99.6                 & 99.9                 & 99.5                 & \multicolumn{1}{c|}{\cellcolor[HTML]{EFEFEF}99.6}                 & 2.1                 & 3.0                 & 1.7                 & 1.3                 & 2.8                 & 2.0                 & 6.1                 & 5.3                 & 27.4                 & 1.6             & 99.2     & \multicolumn{1}{c|}{\cellcolor[HTML]{EFEFEF}13.8}  & \cellcolor[HTML]{FFF8F8}32.2      \\
\multicolumn{1}{l|}{Swin-S(B+J 0.5)}     & \multicolumn{1}{c|}{}                               & 99.6                 & 99.9                 & 99.1                 & \multicolumn{1}{c|}{\cellcolor[HTML]{EFEFEF}99.5} &1.6                 & 2.7                 & 1.6                 & 2.6                 & 4.3                 & 3.8                 & 4.5                 & 3.3                 & 23.4                 & 1.7                 & 99.3                 & \multicolumn{1}{c|}{\cellcolor[HTML]{EFEFEF}13.5}  & \cellcolor[HTML]{FFF8F8}31.9      \\
\multicolumn{1}{l|}{DeiT-S(B+J 0.1)}       & \multicolumn{1}{c|}{}                               & 97.7                 & 99.8                 & 89.9                 & \multicolumn{1}{c|}{\cellcolor[HTML]{EFEFEF}95.8} & 10.5                 & 13.4                 & 11.2                 & 12.4                 & 17.8                 & 15.4                 & 14.9                 & 14.5                 & 36.6                 & 15.9                 & 95.9                 & \multicolumn{1}{c|}{\cellcolor[HTML]{EFEFEF}23.5}  & \cellcolor[HTML]{FFF8F8}38.9      \\
\multicolumn{1}{l|}{DeiT-S(B+J 0.5)}       & \multicolumn{1}{c|}{}                               & 97.1                 & 99.5                 & 87.1                 & \multicolumn{1}{c|}{\cellcolor[HTML]{EFEFEF}94.5} & 11.3                 & 13.6                 & 10.4                 & 13.4                 & 20.7                 & 17.6                 & 15.8                 & 15.2                 & 39.5                 & 18.5                 & 94.6                 & \multicolumn{1}{c|}{\cellcolor[HTML]{EFEFEF}24.6}  & \cellcolor[HTML]{FFF8F8}39.5      \\

\multicolumn{1}{l|}{ResNet50(Fourier)~\cite{Xu-2019-detectfake}}          & \multicolumn{1}{c|}{}    & 99.2                 & 99.9                 & 99.2                 & \multicolumn{1}{c|}{\cellcolor[HTML]{EFEFEF}99.4} & 0.1                 & 0.2                 & 0.1                 & 0.6                 & 0.2                 & 0.1                 & 0.2                 & 0.5                 & 1.0                 & 0.6                 & 0.45                 & \multicolumn{1}{c|}{\cellcolor[HTML]{EFEFEF}0.3} & \cellcolor[HTML]{FFF8F8}21.5     \\
\multicolumn{1}{l|}{Xception(Patch)~\cite{Lucy-2020-understand}}          & \multicolumn{1}{c|}{}         & 53.3                 & 51.8                 & 51.2                 & \multicolumn{1}{c|}{\cellcolor[HTML]{EFEFEF}52.1} & 49.7                 & 48.7                 & 48.5                 & 48.6                 & 51.2                 & 51.6                 & 45.1                 & 50.8                 & 54.7                 & 45.9                 & 49.0                 & \multicolumn{1}{c|}{\cellcolor[HTML]{EFEFEF}49.4}  & \cellcolor[HTML]{FFF8F8}50.0\\ 
\multicolumn{1}{l|}{\deepfake{$F^3$-Net}~\cite{Qian-F3net}}          & \multicolumn{1}{c|}{}         & 91.7                 & 98.8                 & 86.2                 & \multicolumn{1}{c|}{\cellcolor[HTML]{EFEFEF}92.2} & 15.4                 & 13.3                 & 9.5                 & 9.7                 & 17.6                 & 15.3                 & 19.9                 & 21.7                 & 29.4                 & 18.1                 & 97.5                 & \multicolumn{1}{c|}{\cellcolor[HTML]{EFEFEF}24.3}  & \cellcolor[HTML]{FFF8F8}38.8\\ 
\multicolumn{1}{l|}{\deepfake{Gramnet}~\cite{Liu-2020-gramnet}}          & \multicolumn{1}{c|}{}         & 85.1                 & 99.5                 & 82.2                 & \multicolumn{1}{c|}{\cellcolor[HTML]{EFEFEF}88.9} & 14.3                 & 14.6                 & 9.5                 & 13.1                 & 22.0                 & 21.0                 & 19.2                 & 18.4                 & 39.9                 & 27.1                 & 94.3                 & \multicolumn{1}{c|}{\cellcolor[HTML]{EFEFEF}26.6}  & \cellcolor[HTML]{FFF8F8}40.0\\ 
\multicolumn{1}{l|}{\deepfake{ELA-Xception}~\cite{Teddy-2017-elaxception}}          & \multicolumn{1}{c|}{\multirow{-9}{*}{\makecell[c]{\textcolor{MyDarkgray}{Dataset Setting C:}\\ StyleGAN3 (87K)\\StyleGAN3-Train (87K)}}}         & 73.8                 & 99.7                 & 68.6                 & \multicolumn{1}{c|}{\cellcolor[HTML]{EFEFEF}80.7} & 49.1                 & 38.2                 & 35.1                 & 41.4                 & 42.1                 & 41.1                 & 47.3                 & 53.1                 & 73.2                 & 38.5                 & 89.3                 & \multicolumn{1}{c|}{\cellcolor[HTML]{EFEFEF}49.8}  & \cellcolor[HTML]{FFF8F8}56.4\\ \midrule

\multicolumn{1}{l|}{Swin-S(B+J 0.1)}     & \multicolumn{1}{c|}{}                               & 83.2                 & 99.9                 & 99.9                 & \multicolumn{1}{c|}{\cellcolor[HTML]{EFEFEF}94.3}                 & 48.9                 & 92.3                 & 99.9                 & 99.9                 & 99.9                 & 99.9                 & 67.9                 & 95.1                 & 61.4                 & 13.3         &99.3        & \multicolumn{1}{c|}{\cellcolor[HTML]{EFEFEF}79.8}  & \cellcolor[HTML]{FFF8F8}82.9      \\
\multicolumn{1}{l|}{Swin-S(B+J 0.5)}     & \multicolumn{1}{c|}{}                               & 93.5                 & 99.9                 & 99.9                 & \multicolumn{1}{c|}{\cellcolor[HTML]{EFEFEF}97.7} & 47.2                 & 79.9                 & 99.9                 & 99.7                 & 99.8                 & 99.7                 & 59.1                 & 93.5                 & 64.1                 & 10.6                 & 98.8                 & \multicolumn{1}{c|}{\cellcolor[HTML]{EFEFEF}77.4}  & \cellcolor[HTML]{FFF8F8}81.8      \\
\multicolumn{1}{l|}{DeiT-S(B+J 0.1)}       & \multicolumn{1}{c|}{}                               & 98.2                 & 99.9                 & 99.6                 & \multicolumn{1}{c|}{\cellcolor[HTML]{EFEFEF}99.2} & 51.0                 & 69.4                 & 99.8                 & 94.6                 & 98.4                 & 95.0                 & 48.0                 & 56.5                 & 37.3                 & 10.3                 & 96.7                 & \multicolumn{1}{c|}{\cellcolor[HTML]{EFEFEF}68.6}  & \cellcolor[HTML]{FFF8F8}75.3      \\
\multicolumn{1}{l|}{DeiT-S(B+J 0.5)}       & \multicolumn{1}{c|}{}                               & 96.7                 & 99.7                 & 99.0                 & \multicolumn{1}{c|}{\cellcolor[HTML]{EFEFEF}98.4} & 54.8                 & 75.7                 & 99.8                 & 96.0                 & 99.1                 & 97.6                 & 65.7                 & 79.1                 & 55.4                 & 12.1                 & 93.1                 & \multicolumn{1}{c|}{\cellcolor[HTML]{EFEFEF}75.3}  & \cellcolor[HTML]{FFF8F8}80.2      \\

\multicolumn{1}{l|}{ResNet50(Fourier)~\cite{Xu-2019-detectfake}}          & \multicolumn{1}{c|}{} & 58.0                 & 26.8                 & 62.5                 & \multicolumn{1}{c|}{\cellcolor[HTML]{EFEFEF}49.1} & 41.2                 & 51.2                 & 48.4                 & 43.2                 & 50.7                 & 53.8                 & 7.3                 & 53.1                 & 21.4                 & 58.7                 & 64.3& \multicolumn{1}{c|}{\cellcolor[HTML]{EFEFEF}44.8}  & \cellcolor[HTML]{FFF8F8}45.7\\ 
\multicolumn{1}{l|}{Xception(Patch)~\cite{Lucy-2020-understand}}          & \multicolumn{1}{c|}{\multirow{-6}{*}{\makecell[c]{\textcolor{MyDarkgray}{Dataset Setting D:}\\ SD-V1.5Real-dpms-25 (460K)\\IF-V1.0-dpms++-25 (460K)\\StyleGAN3 (87K)\\CC3M-Train (1M)\\StyleGAN3-Train (87K)}}}         & 75.3                 & 69.3                 & 69.8                 & \multicolumn{1}{c|}{\cellcolor[HTML]{EFEFEF}71.4} & 51.9                 & 54.7                 & 50.3                 & 55.5                 & 59.0                 & 58.2                 & 32.3                & 55.3                 & 47.4                 & 43.9                 & 17.0                 & \multicolumn{1}{c|}{\cellcolor[HTML]{EFEFEF}47.7}  & \cellcolor[HTML]{FFF8F8}52.8\\ 

\midrule[1.5pt]

\end{tabular}}
\vspace{-0.5cm}
\label{model_eval_add}
\end{table}
\subsection{More Experiments on MPBench}
We conducted more experiments on MPBench, as shown in Tab.~\ref{model_eval_add}.

\subsection{Evaluate the best model under the same setting used in HPBench.}
\begin{figure*}[!h]
    \centering
    \includegraphics[width=1.0\linewidth]{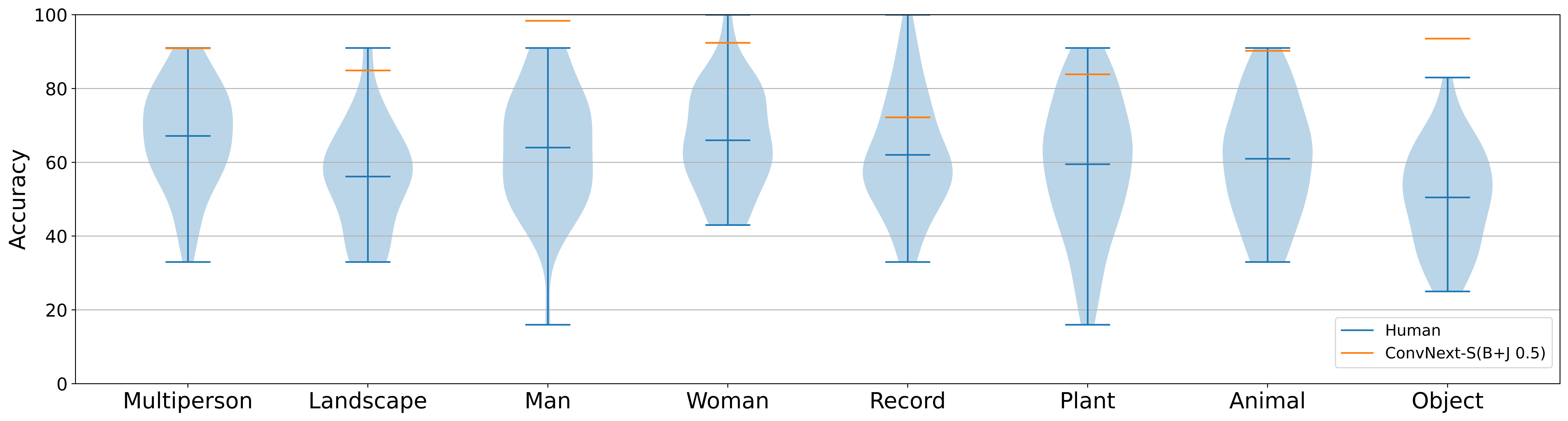}
    \caption{
        \textbf{Human evaluation score distribution and ConvNext-S(B+J 0.5) model score in the same dataset HPBench.}
    }
    \label{All images with different categories for all persons and machine}
\end{figure*}
We also present the results of human and ConvNext-S(B+J 0.5) model on the same dataset HPBench, as shown in Fig.~\ref{All images with different categories for all persons and machine}.
We can find that the results of ConvNext-S are better than human in the results of the two categories: \textit{Man} and \textit{Object}. In the remaining categories, the highest performance of human is better than the performance of the model, but the average performance of human is far worse than the performance of the model.

This demonstrates that it is valuable to study the potential benefits of ensemble the abilities of humans and models in addressing this challenge.

\subsection{Hyperparameters of the Experiments}

Detailed information about the hyperparameters of the experiments in MPBench are shown in Tab.~\ref{convnext_hyper}, Tab.~\ref{swin_hyper}, Tab.~\ref{deit_hyper}, Tab.~\ref{resnet50_hyper} and Tab.~\ref{clipl_hyper}. 

\begin{table}[h]
\centering
\subfloat[{ConvNext-S(B+J 0.1)}]{
\centering
\begin{minipage}{1.0\linewidth}
\begin{center}
\tablestyle{3pt}{1.00}
\begin{tabular}{l|c}
\hline
\textbf{config}  & \textbf{value} \\
\hline
optimizer & AdamW \\
optimizer momentum & {$\beta_1, \beta_2{=}0.9, 0.999$} \\ 
weight decay & {0.05} \\
learning rate & 1e-4 \\
learning rate sch. & cosine decay \\
warmup epochs & 0 \\
epochs& 10 \\
augmentation & {HFlip, RandomResizedCrop(224), GaussianBlur(0.1), JPEG(0.1)} \\
batch size & 1024 \\
dtype & bfloat16 \\
resolution & 224 \\
pretrain & ConvNext-Small-In21k \\
\hline
\end{tabular}
\end{center}
\end{minipage}
}

\centering
\subfloat[{ConvNext-S(B+J 0.5)}]{
\centering
\begin{minipage}{1.0\linewidth}
\begin{center}
\tablestyle{3pt}{1.00}
\begin{tabular}{l|c}
\hline
\textbf{config}  & \textbf{value} \\
\hline
optimizer & AdamW \\
optimizer momentum & {$\beta_1, \beta_2{=}0.9, 0.999$} \\ 
weight decay & {0.05} \\
learning rate & 1e-4 \\
learning rate sch. & cosine decay \\
warmup epochs & 0 \\
epochs& 10 \\
augmentation & {HFlip, RandomResizedCrop(224), GaussianBlur(0.5), JPEG(0.5)} \\
batch size & 1024 \\
dtype & bfloat16 \\
resolution & 224 \\
pretrain & ConvNext-Small-In21k \\
\hline
\end{tabular}
\end{center}
\end{minipage}
}

\vspace{-5pt}
\caption{{\bf Settings for ConvNext-S~\cite{Zhuang-2023-convnext} in MPBench.}}
\label{convnext_hyper}
\end{table}
\newpage
\begin{table}[h]
\centering
\subfloat[{Swin-S(B+J 0.1)}]{
\centering
\begin{minipage}{1.0\linewidth}
\begin{center}
\tablestyle{3pt}{1.00}
\begin{tabular}{l|c}
\hline
\textbf{config}  & \textbf{value} \\
\hline
optimizer & AdamW \\
optimizer momentum & {$\beta_1, \beta_2{=}0.9, 0.999$} \\ 
weight decay & {0.05} \\
learning rate & 1e-4 \\
learning rate sch. & cosine decay \\
warmup epochs & 0 \\
epochs& 10 \\
augmentation & {HFlip, RandomResizedCrop(224), GaussianBlur(0.1), JPEG(0.1)} \\
batch size & 1024 \\
dtype & bfloat16 \\
resolution & 224 \\
pretrain & Swin-Small-In1k \\
\hline
\end{tabular}
\end{center}
\end{minipage}
}

\centering
\subfloat[{Swin-S(B+J 0.5)}]{
\centering
\begin{minipage}{1.0\linewidth}
\begin{center}
\tablestyle{3pt}{1.00}
\begin{tabular}{l|c}
\hline
\textbf{config}  & \textbf{value} \\
\hline
optimizer & AdamW \\
optimizer momentum & {$\beta_1, \beta_2{=}0.9, 0.999$} \\ 
weight decay & {0.05} \\
learning rate & 1e-4 \\
learning rate sch. & cosine decay \\
warmup epochs & 0 \\
epochs& 10 \\
augmentation & {HFlip, RandomResizedCrop(224), GaussianBlur(0.5), JPEG(0.5)} \\
batch size & 1024 \\
dtype & bfloat16 \\
resolution & 224 \\
pretrain & Swin-Small-In1k \\
\hline
\end{tabular}
\end{center}
\end{minipage}
}

\vspace{-5pt}
\caption{{\bf Settings for Swin-S~\cite{Ze-2021-swin} in MPBench.}}
\label{swin_hyper}
\end{table}
\newpage
\begin{table}[h]
\centering
\subfloat[{DeiT-S(B+J 0.1)}]{
\centering
\begin{minipage}{1.0\linewidth}
\begin{center}
\tablestyle{3pt}{1.00}
\begin{tabular}{l|c}
\hline
\textbf{config}  & \textbf{value} \\
\hline
optimizer & AdamW \\
optimizer momentum & {$\beta_1, \beta_2{=}0.9, 0.999$} \\ 
weight decay & {0.05} \\
learning rate & 1e-4 \\
learning rate sch. & cosine decay \\
warmup epochs & 0 \\
epochs& 10 \\
augmentation & {HFlip, RandomResizedCrop(224), GaussianBlur(0.1), JPEG(0.1)} \\
batch size & 1024 \\
dtype & bfloat16 \\
resolution & 224 \\
pretrain & DeiT-Small-In1k \\
\hline
\end{tabular}
\end{center}
\end{minipage}
}

\centering
\subfloat[{DeiT-S(B+J 0.5)}]{
\centering
\begin{minipage}{1.0\linewidth}
\begin{center}
\tablestyle{3pt}{1.00}
\begin{tabular}{l|c}
\hline
\textbf{config}  & \textbf{value} \\
\hline
optimizer & AdamW \\
optimizer momentum & {$\beta_1, \beta_2{=}0.9, 0.999$} \\ 
weight decay & {0.05} \\
learning rate & 1e-4 \\
learning rate sch. & cosine decay \\
warmup epochs & 0 \\
epochs& 10 \\
augmentation & {HFlip, RandomResizedCrop(224), GaussianBlur(0.5), JPEG(0.5)} \\
batch size & 1024 \\
dtype & bfloat16 \\
resolution & 224 \\
pretrain & DeiT-Small-In1k \\
\hline
\end{tabular}
\end{center}
\end{minipage}
}

\vspace{-5pt}
\caption{{\bf Settings for DeiT-S~\cite{Hugo-2021-deit} in MPBench.}}
\label{deit_hyper}
\end{table}
\newpage
\begin{table}[h]
\centering
\subfloat[{ResNet50(B+J 0.1)}]{
\centering
\begin{minipage}{1.0\linewidth}
\begin{center}
\tablestyle{3pt}{1.00}
\begin{tabular}{l|c}
\hline
\textbf{config}  & \textbf{value} \\
\hline
optimizer & SGD \\
optimizer momentum & {$\beta{=}0.9$} \\ 
weight decay & {1e-4} \\
learning rate & 1e-4 \\
learning rate sch. & cosine decay \\
warmup epochs & 0 \\
epochs& 10 \\
augmentation & {HFlip(0.5), RandomResizedCrop(224), GaussianBlur(0.1), JPEG(0.1)} \\
batch size & 512 \\
dtype & bfloat16 \\
resolution & 224 \\
pretrain & ResNet-50-In1k \\
\hline
\end{tabular}
\end{center}
\end{minipage}
}

\centering
\subfloat[{ResNet50(B+J 0.5)}]{
\centering
\begin{minipage}{1.0\linewidth}
\begin{center}
\tablestyle{3pt}{1.00}
\begin{tabular}{l|c}
\hline
\textbf{config}  & \textbf{value} \\
\hline
optimizer & SGD \\
optimizer momentum & {$\beta{=}0.9$} \\ 
weight decay & {1e-4} \\
learning rate & 1e-4 \\
learning rate sch. & cosine decay \\
warmup epochs & 0 \\
epochs& 10 \\
augmentation & {HFlip(0.5), RandomResizedCrop(224), GaussianBlur(0.5), JPEG(0.5)} \\
batch size & 512 \\
dtype & bfloat16 \\
resolution & 224 \\
pretrain & ResNet-50-In1k \\
\hline
\end{tabular}
\end{center}
\end{minipage}
}

\vspace{-5pt}
\caption{{\bf Settings for ResNet50~\cite{Kaiming-2016-resnet} in MPBench.}}
\label{resnet50_hyper}
\end{table}
\newpage
\begin{table}[h]
\centering
\centering
\begin{center}
\tablestyle{3pt}{1.00}
\begin{tabular}{l|c}
\hline
\textbf{config}  & \textbf{value} \\
\hline
optimizer & AdamW \\
optimizer momentum & {$\beta_1, \beta_2{=}0.9, 0.999$} \\ 
weight decay & {0.3} \\
learning rate & 1e-5 \\
learning rate sch. & cosine decay \\
warmup epochs & 0 \\
epochs& 10 \\
augmentation & {HFlip(0.5), RandomResizedCrop(224)} \\
batch size & 512 \\
dtype & bfloat16 \\
resolution & 224 \\
pretrain & openclip-ViT-L-14 \\
\hline
\end{tabular}
\end{center}

\caption{{\bf Settings for CLIP-ViT-L~\cite{Alec-2021-CLIP,Mehdi-2022-openclip} in MPBench.}}
\label{clipl_hyper}
\end{table}

\newpage

\section{Detailed Related Work}

\subsection{Image Generation}
Generating photorealistic images based on given text descriptions has proven to be a challenging task. Previous GAN-based approaches~\cite{Ian-2014-GAN,Andrew-2019-BigGAN,Han-2021-XMCGAN,Tero-2019-stylegan} were only effective within specific domains and datasets, assuming a closed-world setting. However, with the advancements in diffusion models~\cite{Jonathan-2020-DDPM,yang-2021-score}, autoregressive transformers~\cite{vaswani-2017-transformer}, and large-scale language encoders~\cite{Colin-2020-T5,Tom-2020-GPT3,OpenAI-2023-GPT4,Alec-2021-CLIP}, significant progress has been made in high-quality photorealistic text-to-image synthesis with arbitrary text descriptions.

State-of-the-art text-to-image synthesis approaches such as DALL·E 2~\cite{Aditya-2022-DALLE2}, Imagen~\cite{Chitwan-2022-Imagen}, Stable Diffusion~\cite{Robin-2022-sd}, and Midjourney~\cite{midjourney} have demonstrated the possibility of that generating high-quality, photorealistic images with diffusion-based generative models trained on large datasets. Those models have surpassed previous GAN-based models in both fidelity and diversity of generated images, without the instability and mode collapse issues that GANs are prone to.
In addition to diffusion models, other autoregressive models such as Make-A-Scene~\cite{Oran-2022-Make-A-Scene}, CogView~\cite{Ming-2021-CogView}, and Parti~\cite{Jiahui-2022-Parti} have also achieved amazing performance. While diffusion models and autoregressive models exhibit impressive image synthesis ability, they all require time-consuming iterative processes to achieve high-quality image sampling. 
However, the progress made in the field of text-to-image synthesis over the past few years is a testament to the potential of this technology.

\subsection{Deepfake Generation and Detection}


In December 2017, a Reddit user going by the pseudonym "Deepfakes" shared pornographic videos created using open-source AI tools capable of swapping faces in images and videos. Since then, the term "Deepfake" has been widely used to describe the generation of human appearances, particularly facial expressions, through AI methods.
The "Malicious Deep Fake Prohibition Act" of 2018 provides a definition of deepfake as videos and audios that have been realistically but falsely altered and are difficult to identify. Similarly, the "DEEP FAKES Accountability Act" of 2019 defines deepfake as media that is capable of authentically depicting an individual who did not actually participate in the production of the content. Yisroel \textit{et al.}~\cite{Yisroel-2022-DeepfakeSurvey} defines deepfake as believable media generated by a deep neural network.
In essence, deepfake~\cite{Chesney-2019-deepfake} refers to the creation of seemingly realistic but falsified images, audios, videos, and other digital media produced through AI methods, particularly deep learning.

Realistic deepfake media has posed a significant threat to privacy, democracy, national security, and society as a whole. These images and videos have the potential to bypass facial authentication, create political unrest, spread fake news, and even be used for blackmail. The proliferation of fake information through fabricated videos and images can severely undermine our trust in online digital content.
Furthermore, the highly realistic nature of deepfake media makes it difficult for humans to identify them as being falsified. 
Thus, the ability to distinguish between deepfake and real media has become an important, necessary, and urgent matter.

In recent years, there have been many works~\cite{Xu-2019-detectfake,Sheng-2020-GeneratedFake,Utkarsh-2023-UniFake,Lucy-2020-understand,Lakshmanan-2019-Detect,Vishal-2022-Proactive,Davide-2015-Splicebuster,Joel-2020-Frequency,Francesco-2018-Detection} exploring how to distinguish whether an image is AI-generated. 
These works focus on images generated by GANs or small generation models~\cite{Ian-2014-GAN,Andrew-2019-BigGAN,Han-2021-XMCGAN,Tero-2019-stylegan}. 
Due to the limited quality of images generated by those methods, it is easy for humans to distinguish whether a photo is AI-generated or not. 
However, as the quality of generated images continues to improve with the advancement of recent generative models~\cite{Chitwan-2022-Imagen,midjourney,Robin-2022-sd,Aditya-2022-DALLE2}, it has become increasingly difficult for humans to identify whether an image is generated by AI or not. 
Lyu \textit{et al.}~\cite{Yanru-2022-HumanAI} provides an in-depth investigation into communication in human-AI co-creation, specifically focusing on the perceptual analysis of paintings generated by a text-to-image system.
Instead of exploring the human perception of AI-generated paintings, we study the human perception of AI-generated photographic images that may contain contradictions or absurdities that violate reality. Those AI-generated photorealistic images can potentially pose a significant threat to the accuracy of factual information.
In conclusion, the objective of our study is to investigate whether state-of-the-art AI-generated photographic images are capable of deceiving human perception.


\section{Discussion, Broader Impact, Limitation and Conclusion}


\subsection{Discussion}

\paragraph{Can AIGC deceive humans now?}
With the recent rapid advancements in generative AI, AI is now capable of producing highly photorealistic images with rich backgrounds, vivid characters, and beautiful lighting.
Although people may able to occasionally differentiate low-quality AI-generated images, it is becoming more and more difficult to distinguish high-quality AI-generated images from real photography.
In this study, our human evaluation results indicate that the state-of-the-art (SOTA) AI model is able to deceive the human eye to a significant degree (38.7\%).
Moreover, our exploration shows that it is no longer reliable to judge whether an image is real based solely on image quality. 
Instead, people need to consider factors such as over-smoothing portrait faces, coherence, and consistency between objects, and physical laws in the image, which makes the distinguishing process much harder and time-consuming (about 18 seconds for each image in this study).

From another aspect, current AI still can not \textbf{consistently} deceive the human eye.
AI-generated images still have certain defeats which could be used by humans to distinguish fake images. 
Besides, creating such high-quality images requires prompt engineering skills and numerous experiments. 
Even though, a few finely adjusted AI images with misleading information can convey wrong ideas and cause enormous damage.

\paragraph{What the current state-of-the-art image generation model can do and can not do?}
Given suitable prompts, the SOTA image generation model can produce photo-realistic images that are indistinguishable from real photographs, as shown in Fig.~\ref{identify}. 
The prompt can have different formats (e.g., text, image) and arbitrary complexity, including details such as colors, textures, and lighting.
There are lots of potential applications for image generation.
For instance, AIGC can be used to generate images for advertising campaigns, product catalogs, and fashion magazines. 
Since it can easily be controlled by text, AIGC can also be utilized in the film industry to create realistic special effects or even entire scenes, at an extremely low cost.
Furthermore, AIGC can be implemented in the gaming industry to produce immersive and lifelike game worlds.

Although generative AI has impressive image generation capabilities, it currently faces several limitations and challenges, as shown in Fig.~\ref{AIGC defects}. 
One of the most significant challenges is generating images of multiple people with intricate details in a single scene. 
Users can easily infer the authenticity of an image from details.
Furthermore, the current model has difficulty generating realistic human hand gestures and positions, which are crucial for many applications such as sign language recognition and virtual reality. 
In addition, the current state-of-the-art image generation model can produce images with strange details, blurriness, and unrealistic physical phenomena such as lighting issues. These issues limit the model's ability to generate images with high accuracy and fidelity to real-world scenes.
Overall, while the SOTA image generation model has shown remarkable capabilities, it still faces significant challenges that need to be addressed for it to achieve even greater success in the field of image generation.

\subsection{Broader Impact}

\paragraph{Societal risks.}
As AIGC continues to be promoted in various fields, concerns about its societal use have become increasingly prominent. 
These concerns involve various issues such as bias and ethics.
As we have demonstrated, it is getting more and more difficult for humans to distinguish between AI-generated images and real images. 
Therefore, AI models may produce content that contradicts or even absurdly violates reality, posing a serious threat to factual information. 
Photos may then become increasingly difficult to use as evidence in the future, and even serious public opinion effects may result. 
For example, there were many AI-generated images of Trump being arrested on Twitter recently~\cite{bbc2}. 
Such content may be used to spread false information, incite violence, or harm individuals or organizations. 
Besides, AIGC can be used to create realistic virtual characters, which may be used for malicious purposes such as online fraud, scams, or harassment. 

It is crucial for researchers and practitioners in the field of AIGC to develop strategies to mitigate potential negative impacts. This includes developing methods to identify AI-generated images, establishing guidelines for their ethical use, and raising public awareness about their existence and potential impact. Only by working together can we ensure that the benefits of AIGC are fully realized while minimizing its negative consequences.

\paragraph{Positive impacts.}
Given that AI has shown remarkable performance in creating works of art and photography, it is expected to have a significant impact on artists and photographers in the real world. 
People can obtain a large number of desired works or photos at a lower cost, which could compress the market for artists and photographers. 
In this era of fast-food images, where should the new generation of artists and photographers go~\cite{bbc3}? 
However, AI can only generate soulless works, lacking the creativity, imagination, and emotion possessed by human artists and photographers~\cite{Yanru-2022-HumanAI}. 
Even the most advanced AI technology cannot replace the creativity and individuality of human artists and photographers. 
Therefore, although the emergence of AI has indeed brought new challenges and changes to the fields of art and photography, human artists and photographers are still highly valued. 

The emergence of AI technology presents various new opportunities for artists, designers, and users. 
One of the most significant benefits is the ability to create new and innovative visual works, such as digital art and logos, while reducing the time and cost associated with traditional image creation methods. 
AI technology allows people to generate unique and novel images that might not have been possible otherwise, leading to new ideas and inspiration. 
Moreover, AI technology can help optimize existing works of art and photos, leading to improved quality and value. 
For instance, AI can be used to enhance or restore old or damaged images to their original state~\cite{Ziyu-2020-oldphoto}, which can be particularly useful in restoring historic photographs or artworks~\cite{bbc4}. 
AIGC also provides users a more personalized experience by creating images tailored to their personal preferences~\cite{Rinon-2022-textinvision, Nataniel-2022-dreambooth}. 
This customization can lead to more engaging and immersive experiences for users. 

\paragraph{Academic impacts.}
In this study, we conduct a quantitative human evaluation of whether the most advanced AI model can deceive the human eye. 
Results indicate several academic directions that could be explored in the future:

\noindent{\textbf{$\bullet$}}
Since people cannot discern the authenticity of images, a natural question arises: \textit{Can AI distinguish whether an image is generated by AI?} 
Exploring how to use AI to detect AI-generated images is a problem that could be studied ~\cite{Sheng-2020-GeneratedFake}. 
Establishing a detection system to recognize AIGC will greatly ensure the security of society and the credibility of images.

\noindent{\textbf{$\bullet$}}
Even the most advanced image generation model still cannot guarantee the stable generation of high-quality images. 
At the same time, as shown in Fig.~\ref{AIGC defects}, AI-generated images often have certain defects. 
Our failure case analysis will inspire researchers to design better image generation models. 
Exploring how to solve these AIGC defects is an important future research direction.

\noindent{\textbf{$\bullet$}}
There is an interesting phenomenon in MPBench: CLIP-ViT-L (LC)~\cite{Utkarsh-2023-UniFake} freezes the pre-trained backbone and unfreezes the last linear layer. Its generalization in MPBench is very good, but its accuracy in real images has dropped a lot. However, other models initialized from pre-trained models with whole backbone unfrozen have good accuracy in real images, but the generalization in MPBench are not good. This phenomenon shows an interesting research problem: Can we achieve a balance between these two settings? To study how many proportions of backbones should be frozen and how many proportions of backbones should be unfrozen is the best setting for fake image detection task is a good research problem.

\noindent{\textbf{$\bullet$}}
In the real world, it is difficult to obtain comprehensive and diverse data, leading to the famous problem of data imbalance~\cite{Justin-2019-imbalance}. 
Using imbalanced data will result in various issues such as the long-tail problem~\cite{Yifan-2021-longtail} and bias problem~\cite{Ninareh-2022-Bias}. 
Since the current state-of-the-art image generation model can already produce high-quality data, exploring how to use the image generation model to solve these problems and test the current model's robustness and bias is a problem that could be studied.

\subsection{Limitation}
While this work has so far provided several state-of-the-art and large-scale training and validation datasets, as well as several powerful benchmarks, this section explores the limitations of the  which are expected to be addressed in future studies.

\paragraph{Dataset limitation.}
Our training dataset Fake2M only includes three advanced models: Stable Diffusion v1.5~\cite{Robin-2022-sd}, IF~\cite{if}, and StyleGAN3~\cite{Tero-2020-stylegan3}, limited by the absence of open-source and powerful open vocabulary GAN~\cite{kang-2023-gigagan} and Autoregressive models~\cite{Jiahui-2022-Parti}.
Due to the lack of API, we are unable to provide a training dataset for Midjourney V5.
We hope that future work can further improve the diversity and size of the training dataset to include more powerful generative models.

For the validation datasets, we only include validation datasets for the most advanced generative models, without including validation datasets for other tasks, such as deepfake and low-level tasks.
We hope that future work can further improve the diversity of the validation dataset to include more tasks about fake images.

\paragraph{Benchmark limitation.}
Due to the resource limitations, our high-quality human evaluation HPBench only recruits 50 participants.
Our human evaluation also lacks diversity in terms of participant background, as it only includes a few attributes such as age, AIGC-background and gender.
We hope that future work can further improve the diversity and size of the participants.


\subsection{Conclusion}
In this study, we present a comprehensive evaluation of both human discernment and contemporary AI algorithms in detecting fake images. 
Our findings reveal that humans can be significantly deceived by current cutting-edge image generation models: high-quality AI-generated images can be comparable to real photographs.
In contrast, AI fake image detection algorithms demonstrate a superior ability to distinguish authentic images from fakes.
Despite this, our research highlights that existing AI algorithms, with a considerable misclassification rate of 13\%, still face significant challenges. We anticipate that our proposed dataset, \textbf{Fake2M}, and our dual benchmarks, \textbf{HPBench} and \textbf{MPBench}, will invigorate further research in this area and assist researchers in crafting secure and reliable AI-generated content systems.
As we advance in this technological era, it is crucial to prioritize responsible creation and application of generative AI to ensure its benefits are harnessed positively for society.

We have focused on the surprising abilities of the current SOTA image generation model, but we have not addressed the core questions of why and how it achieves such remarkable intelligence, nor the most important issues of how to ensure the security and credibility of AIGC images.
It is a significant challenge for researchers to develop secure and reliable AIGC systems that can be trusted for various real-world applications, and ensure the responsible and ethical use of AIGC technology in the future.
It is time to prioritize responsible development and the use of generative AI to ensure a positive impact on society.



\newpage
\newpage
\newpage
\newpage



\setcitestyle{square,numbers,comma}

\end{document}